\pdfoutput=1

\PassOptionsToPackage{table}{xcolor}
\documentclass[11pt]{article}
\usepackage[final]{EMNLP2023}
\usepackage{times}
\usepackage{latexsym}
\usepackage[T1]{fontenc}
\usepackage[utf8]{inputenc}
\usepackage{microtype}
\usepackage{inconsolata}
\usepackage{graphicx}
\usepackage{placeins}
\usepackage{enumitem}
\usepackage{morefloats}
\usepackage[table]{xcolor}
\usepackage{comment}
\usepackage{listings}
\definecolor{blue}{HTML}{AADDFF}
\definecolor{green}{HTML}{AAFFAA}
\definecolor{yellow}{HTML}{FFFF88}
\definecolor{red}{HTML}{FF8888}
\definecolor{orange}{HTML}{FFBB55}
\definecolor{purple}{HTML}{AAAAFF}
\definecolor{purple2}{HTML}{CDC9FF}
\definecolor{pink}{HTML}{FFAAFF}
\definecolor{grey}{HTML}{CCCCCC}
\newcommand\blfootnote[1]{%
  \begingroup
  \renewcommand\thefootnote{}\footnote{#1}%
  \addtocounter{footnote}{-1}%
  \endgroup
}
\title{Uncovering Cultural Representation Disparities in Vision-Language Models}
\author{
 \textbf{Ram Mohan Rao Kadiyala*              \textsuperscript{1,2}},
 \textbf{Siddhant Gupta*                      \textsuperscript{2,3}},
 \textbf{Jebish Purbey*                       \textsuperscript{2}},\\
 \textbf{Srishti Yadav                        \textsuperscript{4}},
 \textbf{Suman Debnath                        \textsuperscript{6}}
 \textbf{Alejandro Salamanca                  \textsuperscript{5}},
 \textbf{Desmond Elliott                      \textsuperscript{4}},
 \\ \\ 
 \textsuperscript{1}Traversaal.ai
 \textsuperscript{2}Cohere Labs Community
 \textsuperscript{3}IIT Roorkee\\
 \textsuperscript{4}University of Copenhagen
 \textsuperscript{5}Cohere Labs
 \textsuperscript{6}Amazon
 \\ \\
 \small{\textbf{Datasets \& Code:} \href{https://huggingface.co/datasets/Biases/CulturalBiases-2025}{Datasets \& Code}}\\
}
\begin{document}
\maketitle
\blfootnote{$^\ast$ Equal Contribution.}
%%%%%%%%%%%%%%%%%%%%%%%%%%%%%%%%%%%%%%%%%%%%%%%%%%%%%%%%%%%%%%%%%%%%%%%%%%%%%%%%%%%%%%%%%%%%%%%%%%%%%%%%%%%%
\begin{abstract}
Vision-Language Models (VLMs) have demonstrated impressive capabilities across a range of tasks, yet concerns about their potential biases exist. This work investigates the extent to which prominent VLMs exhibit cultural biases by evaluating their performance on an image-based country identification task at a country level. Utilising the geographically diverse Country211 \citep{country211} dataset, we probe several VLMs under various prompting strategies: open-ended questions, multiple-choice questions (MCQs), including challenging setups like multilingual and adversarial settings. Our analysis aims to uncover disparities in model accuracy across different countries and question formats, providing insights into how training data distribution and evaluation methodologies might influence cultural biases in VLMs. The findings highlight significant variations in performance, suggesting that while VLMs possess considerable visual understanding, they inherit biases from their pre-training data and scale that impact their ability to generalize uniformly across diverse global contexts.
\end{abstract}
%%%%%%%%%%%%%%%%%%%%%%%%%%%%%%%%%%%%%%%%%%%%%%%%%%%%%%%%%%%%%%%%%%%%%%%%%%%%%%%%%%%%%%%%%%%%%%%%%%%%%%%%%%%%
\section{Introduction}

VLMs have rapidly advanced, demonstrating exceptional capabilities in integrating visual and textual information for a wide array of tasks, from image captioning to visual question answering \citep{llava,flamengo,cogvlm}. These models are increasingly being deployed in diverse applications, impacting areas such as education, healthcare, and public services globally \citep{vlmforvision}. 

\begin{figure*}
    \centering
    \includegraphics[width=0.79\linewidth]{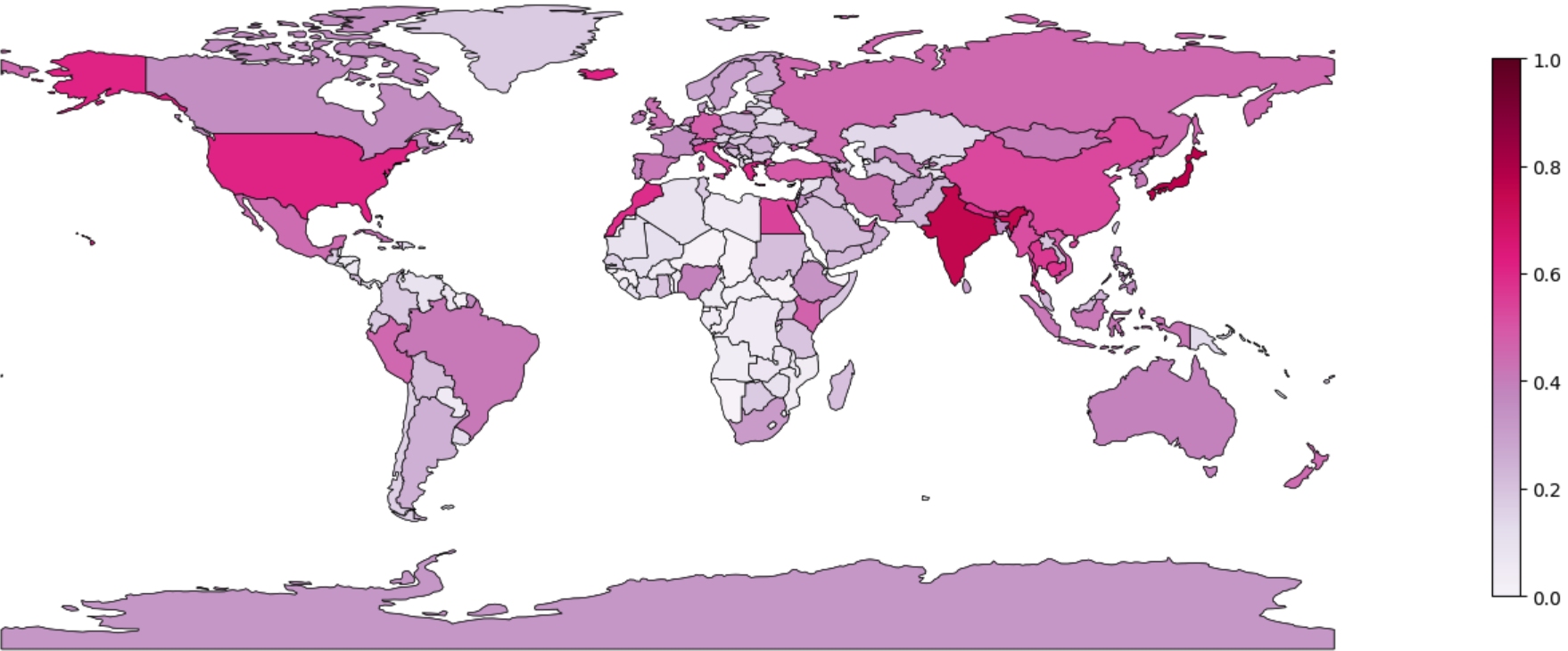}
    \caption{Visualization of the average country-wise recognition accuracy across the VLMs studied in this paper. VLMs perform well at recognizing images from North American and Western European countries, but there are clear disparities in performance for African and Central American countries.}
    \label{figure:1}
\end{figure*}

%%%------------------------------------------------------------------------------------------------------%%%
However, as their influence grows, so do concerns regarding their potential to perpetuate and even amplify societal biases present in their training data \citep{menalsoshop,vlmbias,imagesspeak}. Cultural and geographical biases are of particular concern because they can lead to unequal performance and representation across different populations and regions of the world \citep{culturealignment,llmsgeobiased}. Defining "culture" is inherently complex, encompassing a broad spectrum of social norms, values, practices, languages, and historical contexts that shape the lived experiences of individuals and communities \citep{Kroeber_Untereiner_Kluckhohn_1985}. Establishing culture in computational settings presents a persistent challenge due to its multifaceted and dynamic nature. Empirical studies employ tractable proxies such as demographic or geographic proxies to enable systematic analysis \citep{towards_measuring_culture, imagevalues}. While nation-level aggregation can mask sub-national heterogeneity, prior work in human–computer interaction and cultural analytics has demonstrated that country labels often serve as a practical proxy for coarse‐grained cultural signals when large‐scale analyses are required \citep{obradovich2022expanding}. 

In order to quantify cultural disparities in VLMs, we adopt image-based country identification as a concrete proxy task in which a model must infer an image's country of origin solely from visual cues, while also providing a justification. Prior work has shown that geolocation tasks reveal representational imbalances in visual models, as performance often correlates with the prevalence of training data from different regions \citep{Cultural_and_socioeconomic_diversity}.

The main contributions of this paper are:

\begin{enumerate}
    \item We introduce a scalable framework to evaluate cultural biases in VLMs using an image-based country identification task over 211 countries, leveraging the geographically diverse and balanced Country211 dataset.
    \item We systematically probe VLMs under varied settings—open-ended and multiple-choice questions (MCQs) with both random and culturally similar distractors—alongside multilingual prompts in five languages, to capture nuanced cultural and linguistic disparities. \footnote{Due to cultural similarities, misclassification among similar countries is more likely than misclassification with an unrelated country. MCQ with random and similar distractors tested the VLMs in both scenarios as to whether misclassification would occur when all distractors are neither neighboring nor similar countries}
    \item We examine model robustness to image perturbations and analyse performance across nine image categories (e.g. architecture, landscape, food etc), revealing the influence of image content on cultural bias.
    \item Our findings show that VLM biases do not consistently favour Western countries; instead, biases often reflect over representation of certain popular countries (e.g., India, USA) in the training data \footnote{For deliberately under specified inputs without country names, the generated images most reflect the surroundings of the United States followed by India. \citep{basu2023inspectinggeographicalrepresentativenessimages}}, suggesting a more complex bias landscape.

\end{enumerate}%%%%%%%%%%%%%%%%%%%%%%%%%%%%%%%%%%%%%%%%%%%%%%%%%%%%%%%%%%%%%%%%%%%%%%%%%%%%%%%%%%%%%%%%%%%%%%%%%%%%%%%%%%%%
\section{Related Works}
\begin{table*}
    \centering
    \resizebox{\textwidth}{!}{\
    \begin{tabular}{ccccccc}
    \noalign{\hrule height 1pt}
    \rowcolor{purple2}\textbf{\small{Prior Work}} & \textbf{\small{Eval Method}} & \textbf{\small{Multilingual?}} & \textbf{\small{Adversarial?}} & \textbf{\small{Categories}} & \textbf{\small{Total Sample Count}} & \textbf{\small{Domain}} \\
    \noalign{\hrule height 1pt}
    \small{CulturalVQA \citep{culturalvqa}} & Open-Ended & No & No & 11 Countries & 2,328 & 5 Categories \\
    \small{WorldCuisines \citep{winata2025worldcuisinesmassivescalebenchmarkmultilingual}} & Both & Yes (30 languages) & Yes & 189 Countries & 6,045 & Only Food \\
    \small{Food-500 CAP \citep{food500cap}} & Open-Ended & No & Yes & 7 Regions & 24,700 & Only Food \\
    \small{MOSAIC-1.5k \citep{10924504}} & Open-Ended & No & No & N/A & 1,500 & 3 Categories \\
    \small{See It From My Perspective \citep{See_it_from_My_Perspective}} & Open-Ended & Yes (2 languages) & No & 2 Regions & 38,479 & 4 Categories \\
    \small{CVQA \citep{CVQA}} & MCQ & Yes (31 languages) & Yes & 39 Countries & 5,239 & 10 Categories \\
    \small{GIMMICK \citep{schneider-etal-gimmick-2025}} & Both(MCQ)** & No & No & 144 Countries & 7,239(1,741)** & - \\
    \noalign{\hrule height 1pt}
    \textbf{Ours} & \textbf{Both} & Yes (5 languages) & \textbf{Yes} & \textbf{211 Countries} & \textbf{63,300} & \textbf{9 Categories} \\    
    \noalign{\hrule height 1pt}
    \end{tabular}
    }
    \caption{Overview of prior datasets used in cultural recognition experiments.}
    \caption*{**The vales in brackets indicate the features in the Country recognition task subset, while the first values indicate that of the whole dataset.}
    \label{table:1}
\end{table*}

Recent work has increasingly explored the socio-cultural dimensions of Large Language Models (LLMs), including how they encode, express, and respond to culturally specific knowledge. Studies have examined value alignment \citep{choenni2024self}, moral reasoning across languages \citep{agarwal-etal-2024-ethical}, and cultural persona \citep{culturealignment}, while also uncovering strong Western biases in model outputs \citep{naous-etal-2024-beer} which risk marginalizing cultural diversity if deployed in real world. There have also been efforts to address these concerns, like prompting based on ethnographic fieldwork \citep{culturealignment} and fine-tuning culture-specific LLMs \citep{culturellm}. Similar studies have been extended for Vision Language Models (VLMs) starting from \citep{liu-etal-2021-visually} over cultural aspects, but in a weaker capacity \citep{BridgetheDivide} showed that CLIP \citep{open-clip} struggled in data for poor socio-economic groups worldwide in the Dollar Street dataset \citep{DollarStreet}. State-of-the-art off-shelf VLMs score much higher on images depicting Western scenes than equivalent East-Asian scenes for every vision task, such as identification, question-answering, and art emotion classification \citep{See_it_from_My_Perspective}. Similarly, \citep{liu2025culturevlmcharacterizingimprovingcultural,imagevalues} reveals that VLMs show stronger performance in Western concepts and weaker results in African and Asian contexts. These findings align with the fact that large pretraining corpora are dominated by high-resource languages and regions. Of the samples that can be geo-located in the OpenImages dataset \cite{OpenImages}, 32\% were from only the United States, and 60\% came from only six Western countries \citep{opendataset_representation}. Such imbalances translate into a “Western bias” in model behavior \citep{Vries_2019_CVPR_Workshops}. 

\paragraph{Datasets \& Benchmarks :}
\begin{figure*}
    \centering
    \includegraphics[width=0.80\linewidth]{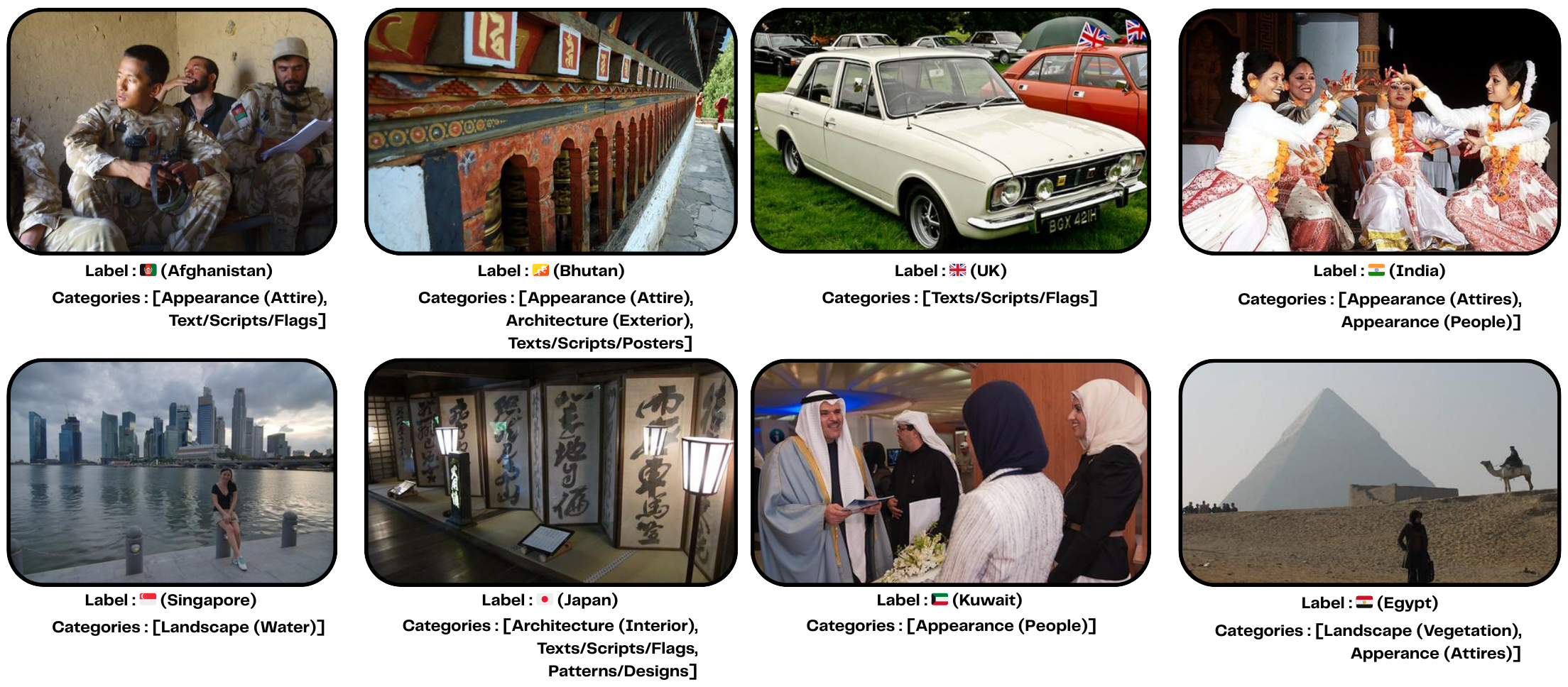}
    \caption{Examples of the Country211 dataset, alongside automatically-predicted categories for each image, showcasing the visual diversity of the examples to be classified.}
    \label{figure:11}
\end{figure*}

To probe these biases, a growing body of work has constructed specialized datasets and benchmarks with cross-cultural content, such as MOSAIC-1.5k \citep{10924504}, CULTURAL-VQA \citep{culturalvqa}, and GlobalRG \citep{GlobalRG}. Many works also opt for probing specific aspects of culture, such as food \citep{li-etal-2024-foodieqa}, race \citep{huang2025visbiasmeasuringexplicitimplicit}, art \citep{mohamed-etal-2024-culture}, etc., instead of providing an overall view for bias study. \citep{winata2025worldcuisinesmassivescalebenchmarkmultilingual} introduced WorldCuisines for Food Vision Question Answering and country identification and found that VLMs often fail on adversarially misleading contexts or less-common cuisines. \citep{food500cap} introduced the Food-500 CAP dataset and observed that most models exhibited geographical culinary biases. Several studies have also treated country-of-origin or geolocation as a proxy for cultural provenance. WorldCuisines includes a country identification task to reveal failures on uncommon or misleading contexts \citep{winata2025worldcuisinesmassivescalebenchmarkmultilingual}, and Food-500 CAP finds systematic mismatches between predicted and actual countries of culinary images \citep{food500cap}. Even in datasets like Dollar Street \citep{DollarStreet} or OpenImages \citep{OpenImages}, geographic metadata has been used to analyze representational imbalances across regions \citep{BridgetheDivide, opendataset_representation}, demonstrating that country-level annotations provide a practical signal for probing cultural and geographic bias in VLMs. 

\paragraph{Impact of Evaluation:} The format of evaluation also impacts bias measurement. Many of the above benchmarks use multiple-choice or binary questions, which can mask a model’s true understanding. Since language choice can influence bias, benchmarks are often performed across multiple languages. \citep{CVQA} showed that the performance of LLaVA-1.5-7B dropped by 19.6\% when prompted without multiple choices for CVQA. Models also showed lower performance when prompted in native language of the image's country of origin. However, \citep{See_it_from_My_Perspective} observed that prompting in a culturally closer language can reduce Western bias in some VLMs. It was also observed that people of different cultures are capable of differently capable of describing what they see in an image \citep{van-miltenburg-etal-2017-cross}. We build on these insights by comparing open-ended vs. multiple-choice prompts (including “hard” questions with challenging distractors) and by evaluating in both English and native languages, to see how the prompting strategy affects cultural bias in VLMs.

%%%%%%%%%%%%%%%%%%%%%%%%%%%%%%%%%%%%%%%%%%%%%%%%%%%%%%%%%%%%%%%%%%%%%%%%%%%%%%%%%%%%%%%%%%%%%%%%%%%%%%%%%%%%
\section{Dataset Used}
The primary dataset used for the experiments is the Country211 \citep{open-clip} dataset which was a subset of images from YFCC100M \citep{thomee2016yfcc100m} having GPS coordinates associated with them. The images cover several domains including but not limited to - exterior architecture, interior architecture, landscape (vegetation, nature, sky view), people's appearance, attires, scripts, texts, posters, etc. The GPS coordinates associated with the images were then used to map them to individual countries. ISO-3166 \footnote{\url{https://en.wikipedia.org/wiki/List_of_ISO_3166_country_codes}} codes representing each country were used as labels for each image. ISO labels were used for consistency as country names used by the VLMs were not deterministic i.e Britain was also used simultaneously in place of Great Britain or UK or its constituents, proving the list of tags and corresponding country names led to the models responding consistently with no observable difference in performance. For our experiments, we utilized this dataset, which consists of 21.1 K images, i.e 100 images each from 211 countries.
\paragraph{Key Differences:} Existing benchmarks highlight cultural blind spots in VLMs, but they generally either cover fewer categories or countries or are restricted to specialized domains. Our work differs by using an image-based country-identification task over 211 countries, providing much broader geographic coverage and adversarial probing. Further, the datasets utilized in the prior works utilize a images that might be easier to classify, including but not limited to close up shots of food items, popular monuments being the primary object in an image etc. The dataset we utilized introduces a lot of noise and randomness in a majority of images as seen in \autoref{figure:11}\footnote{The images were part of OpenAI's YFCC100M and come with pre-verified country labels. Although some samples might be difficult to classify even for a native, The primary goal was to uncover cultural biases using the features the VLMs could probably misclassify it with a culturally similar or neighboring country, but frequently misclassify it with a very dissimilar country.}. For instance, the examples shown from UK, India and Egypt might be easy to classify, but the examples from Afghanistan and Kuwait require grasping certain features and their associated knowledge i.e the headgear pattern of the Kuwait image and how it is different from other countries in the region. Th example from Afghanistan requires noticing the afghan flag, while the appearance of the person in the left may try to mislead the VLM due to an appearance of different ethnicity.

%%%------------------------------------------------------------------------------------------------------%%%

\begin{figure}
    \centering
    \includegraphics[width=0.74\linewidth]{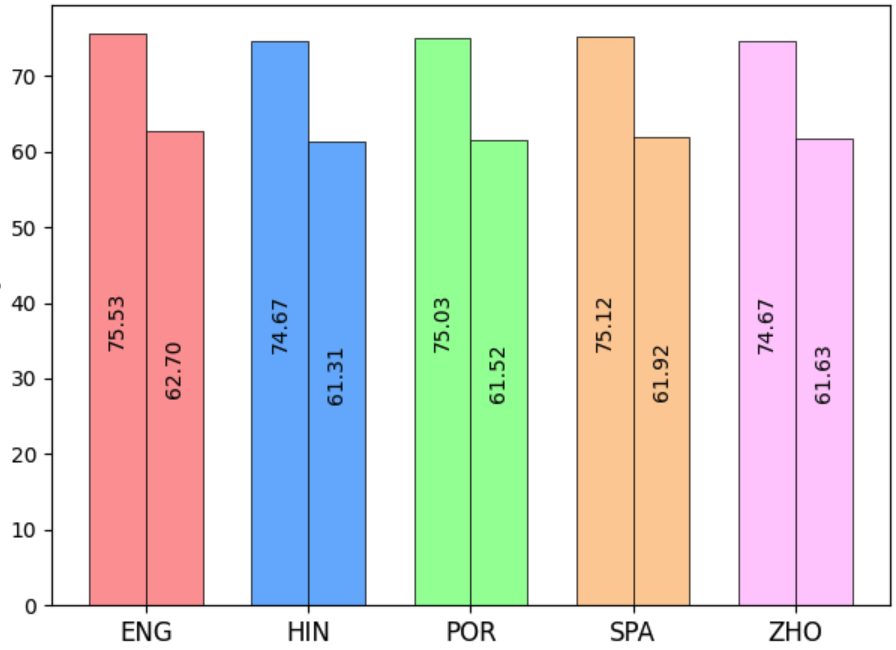}
    \caption{Model-wise averaged accuracy when varying the prompt language or selection of MCQA alternatives (left: random; right: similar). Performance is consistent across conditions.}
    \label{figure:6}
\end{figure}
%%%%%%%%%%%%%%%%%%%%%%%%%%%%%%%%%%%%%%%%%%%%%%%%%%%%%%%%%%%%%%%%%%%%%%%%%%%%%%%%%%%%%%%%%%%%%%%%%%%%%%%%%%%%
\section{Experiments}
\textbf{Prompt Variations:} We probed each VLM under three complementary prompting paradigms. 
\begin{enumerate} [itemsep=-0.5ex] % Adjust itemsep to control vertical spacing
    \item open-ended questions
    \item MCQs (with random distractors)
    \item MCQs (with similar distractors)
\end{enumerate}
\textbf{Image perturbations:} Open-ended experiments were done with these adversarial changes: 

\begin{enumerate} [itemsep=-0.5ex] % Adjust itemsep to control vertical spacing
    \item Rotation by 90$^{\circ}$ clockwise,
    \item Rotation by  90$^{\circ}$  anti-clockwise
    \item Flipping the image
    \item  Gray-scaling the image
\end{enumerate}

However, open ended experiments pose challenges for objective scoring due to semantic variability. Second, multiple-choice questions (MCQs) with random distractors yield correctness metrics yet may understate subtle biases if distractors are easily ruled out. Third, challenging MCQs with similar distractors force models to discriminate between culturally proximate options, thus exposing fine-grained bias patterns. The MCQs are designed as part of discriminative probing and to assess the disparity in the model's cultural knowledge. 
\\ \\
\textbf{Linguistic Variations : }We further extend discriminative proving to a multilingual setting, prompting models in five languages : (English, Hindi, Chinese, Portuguese, Spanish) to assess the intersection of cultural and linguistic biases. 

\textbf{Model Variations : }A diverse set of VLMs were tested including both proprietary and open-weight models of varying sizes: Gemini-2.5-Flash, Gemma-3-27B \citep{gemmateam2025gemma3technicalreport}, Aya-Vision-8B, Aya-Vision-32B \citep{dash2025ayavisionadvancingfrontier}, GPT-4o-Mini \citep{openai2024gpt4technicalreport}, \citep{geminiteam2025geminifamilyhighlycapable}.

The experiments being repeated with each permutation of features lead to a total of 168.8 K samples tested. Inference was done in JSON format with the default hyperparameters for each of the models tested through Cohere \footnote{\url{https://docs.cohere.com/cohere-documentation}} and OpenRouter's API \footnote{\url{https://openrouter.ai/docs/quickstart}}. More on the JSON formatting and prompts used can be found in \autoref{sec:appendix-D}.
%%%%%%%%%%%%%%%%%%%%%%%%%%%%%%%%%%%%%%%%%%%%%%%%%%%%%%%%%%%%%%%%%%%%%%%%%%%%%%%%%%%%%%%%%%%%%%%%%%%%%%%%%%%%
\subsection{Open-Ended Evaluation}
For the open-ended experiments, we asked each model to provide information on 4 areas: (1) name of the country, (2) country selection rationale in a few sentences, (3) a score from 0 to 100 representing the confidence in the classification, (4) and up to 6 features from the image as a list that had influence in the decision. The accuracies of each country obtained using each of the VLMs used can be seen in \autoref{figure:3}. The accuracies of many countries were far lower especially in Eastern Europe, South America, Africa and Central Asia. This gap between country level accuracies was far higher in open ended experiments compared to the multiple-choice experiments .
%%%%%%%%%%%%%%%%%%%%%%%%%%%%%%%%%%%%%%%%%%%%%%%%%%%%%%%%%%%%%%%%%%%%%%%%%%%%%%%%%%%%%%%%%%%%%%%%%%%%%%%%%%%%
\subsection{Evaluation through random distractors in multiple-choice questions}
For these experiments, we asked each model to provide information on 4 areas: (1) name of the country, (2) label of the chosen country from the choices provided (3) country selection rationale in a few sentences, and (4) a score from 0 to 100 representing the confidence in the classification. For these experiments, 4 countries were chosen at random from among the other 210 countries for each sample as distractors. The order of options were then shuffled such that the distribution of correct answer's location is made uniform. Compared to other settings, this setting led to the highest average of accuracies obtained due to the clearly contrasting nature of distractors used. However, many central African nations still face a recognition bias likely due to low representation in training data. This was observed across all VLMs that were tested as seen in \autoref{figure:4}. 
%%%%%%%%%%%%%%%%%%%%%%%%%%%%%%%%%%%%%%%%%%%%%%%%%%%%%%%%%%%%%%%%%%%%%%%%%%%%%%%%%%%%%%%%%%%%%%%%%%%%%%%%%%%%
\subsection{Evaluation through similar distractors in multiple-choice questions}
Similar to the prior experiments with MCQs using random distractors, in this setting use similar nations as distractors. These were chosen from among the bordering nations. Any countries with high similarity in culture if any were added manually. (Ex : Spain -> Mexico). This led to the average of accuracies dropping considerably due the challenging nature of the options presented to the models. However, the drops were observed for only a few countries where choosing similar distractors led to these countries' images being classified as belonging to one of their popular neighbors. This can be observed in \autoref{figure:4} and \autoref{figure:5}.
%%%%%%%%%%%%%%%%%%%%%%%%%%%%%%%%%%%%%%%%%%%%%%%%%%%%%%%%%%%%%%%%%%%%%%%%%%%%%%%%%%%%%%%%%%%%%%%%%%%%%%%%%%%%
\section{Results}
The results for experimental setting over countries of each region can be seen in \autoref{table:2}. Additionally \autoref{table:2001} demonstrates the statistical significance of the results which presents Pearson's $\chi^2$ and Cochran's Q test results for every evaluation condition and its subcategories. The $\chi^2$ p-values of 0.0 (effectively under flowing) reveal that the six VLMs differ in accuracy in a highly significant way, confirming that VLM biases drive overall performance differences. In contrast, the Cochran's Q p-values of 1.0 indicate no significant variation among sub-conditions (whether comparing original, rotated, or grayscale images, or across the five languages), showing that these perturbations do not meaningfully alter each VLM's overall accuracy, but do cause changes in country wise accuracy as seen in \autoref{figure:7} and \autoref{figure:8}.

\begin{table}[ht]
    \centering
    \begin{tabular}{lrrrr}
        \noalign{\hrule height 1pt}
        \rowcolor{purple2} \small{Condition} & $\chi^2$ & $p$ & $Q$ & $p_Q$ \\
        \noalign{\hrule height 1pt}
        \small{Open-ended} &&&& \\
        \noalign{\hrule height 1pt}
        \small{Original}  & 5851.81 & \tiny{$\sim0.0$} & -0.727 & \tiny{$\sim1.0$} \\
        \small{Rotated}   & 4666.49 & \tiny{$\sim0.0$} & -0.952 & \tiny{$\sim1.0$} \\
        \small{Greyscale} & 4280.08 & \tiny{$\sim0.0$} & -0.725 & \tiny{$\sim1.0$} \\
        \noalign{\hrule height 1pt}
        \small{MCQ-S} &&&& \\
        \noalign{\hrule height 1pt}
        \small{Overall}   & {26274.855} & \tiny{$\sim0.0$} & {-0.151} & {\tiny{$\sim1.0$}} \\
        \small{ENG}       & 5232.00   & \tiny{$\sim0.0$} & -0.145 & \tiny{$\sim1.0$} \\
        \small{HIN}       & 5197.50   & \tiny{$\sim0.0$} & -0.153 & \tiny{$\sim1.0$} \\
        \small{POR}       & 5309.00   & \tiny{$\sim0.0$} & -0.155 & \tiny{$\sim1.0$} \\
        \small{SPA}       & 5256.72   & \tiny{$\sim0.0$} & -0.151 & \tiny{$\sim1.0$} \\
        \small{ZHO}       & 5345.21   & \tiny{$\sim0.0$} & -0.155 & \tiny{$\sim1.0$} \\
        \noalign{\hrule height 1pt}
        \small{MCQ-R} &&&& \\
        \noalign{\hrule height 1pt}
        \small{Overall}   & 23855.61 &\tiny{$\sim0.0$} & -0.074 & \tiny{$\sim1.0$} \\
        \small{ENG}       & 4835.90  &\tiny{$\sim0.0$} & -0.073 & \tiny{$\sim1.0$} \\
        \small{HIN}       & 4584.62  &\tiny{$\sim0.0$} & -0.073 & \tiny{$\sim1.0$} \\
        \small{POR}       & 4873.45  &\tiny{$\sim0.0$} & -0.076 & \tiny{$\sim1.0$} \\
        \small{SPA}       & 4782.98  &\tiny{$\sim0.0$} & -0.074 & \tiny{$\sim1.0$} \\
        \small{ZHO}       & 4801.08  &\tiny{$\sim0.0$} & -0.076 & \tiny{$\sim1.0$} \\
        \noalign{\hrule height 1pt}
    \end{tabular}
    \caption{Statistical tests for each evaluation condition and subcategory : Open ended, MCQ-Random (\textbf{MCQ-R}), MCQ-Similar (\textbf{MCQ-S})}
    \label{table:2001}
\end{table}
%%%------------------------------------------------------------------------------------------------------%%%
\begin{table}
    \centering
    \begin{tabular}{cccc}
    \noalign{\hrule height 1pt}
    \rowcolor{purple2} & & \multicolumn{2}{c}{\textbf{\small{MCQA}}} \\
    \rowcolor{purple2}\textbf{\small{Region}} & \textbf{\small{Open-Ended}} & \textbf{\small{Similar}} & \textbf{\small{Random}} \\
    \noalign{\hrule height 1pt}
    \small{North America}               & \small{41.9} & \small{73.7} & \small{80.2} \\
    \small{Central America}             & \small{11.1} & \small{69.7} & \small{68.0} \\
    \small{Caribbean}                & \small{13.6} & \small{50.5} & \small{71.4} \\
    \small{South America}               & \small{20.4} & \small{70.9} & \small{68.7} \\
    \small{Oceania}                  & \small{19.0} & \small{57.5} & \small{68.9} \\
    \small{Western Europe}              & \small{30.9} & \small{57.9} & \small{77.5} \\
    \small{Northern Europe}             & \small{25.3} & \small{60.6} & \small{79.4} \\
    \small{Eastern Europe}              & \small{26.6} & \small{53.4} & \small{75.9} \\
    \small{Middle East}             & \small{29.3} & \small{68.4} & \small{77.1} \\
    \small{Central Asia}                    & \small{26.7} & \small{53.5} & \small{78.1} \\
    \small{East Asia}               & \small{43.6} & \small{71.6} & \small{83.8} \\
    \small{Southeast Asia}              & \small{41.7} & \small{67.5} & \small{81.7} \\
    \small{South Asia}              & \small{49.1} & \small{69.0} & \small{85.5} \\
    \small{North Africa}                    & \small{31.9} & \small{54.3} & \small{78.9} \\
    \small{Central Africa}              & \small{11.8} & \small{57.0} & \small{68.2} \\
    \small{Southern Africa}          & \small{20.4} & \small{74.2} & \small{74.2} \\
    \noalign{\hrule height 1pt}
    \textbf{\small{Overall}}         & \small{27.7} & \small{63.1} & \small{76.1} \\
    \noalign{\hrule height 1pt}
    \end{tabular}
    \caption{Region-wise averaged accuracy across models. There are consistent disparities in performance across different regions, regardless of the prompting method.}
    \label{table:2}
\end{table}
%%%%%%%%%%%%%%%%%%%%%%%%%%%%%%%%%%%%%%%%%%%%%%%%%%%%%%%%%%%%%%%%%%%%%%%%%%%%%%%%%%%%%%%%%%%%%%%%%%%%%%%%%%%%
\subsection{Effect of Language of Inputs on Results}
The average of country level accuracies compared to each language as input can be seen in \autoref{figure:6}. The language used for inputs had a very little effect i.e <2\% for all languages. But at a country level, most countries remained unaffected by language of the prompt to a large extent with change in accuracy less that 0.1\%. The only cases with a noticeable change in accuracy are some but not all of the countries that speak the target language predominantly. For example, Changing the input language from English to Spanish improved accuracy for Spain but the change over Latin-American countries was negligible. Similarly, while switching to Portuguese had improved the accuracy for Brazil, it lead to a drop in accuracy for Portugal. Overall, the input language improves performance for some countries primarily associated with the language used. The results also partially contradict prior findings that prompting in culturally similar languages reduces western bias \citep{See_it_from_My_Perspective}. 
%%%------------------------------------------------------------------------------------------------------%%%
\begin{figure}[!ht]
    \centering
    \includegraphics[width=0.9\linewidth]{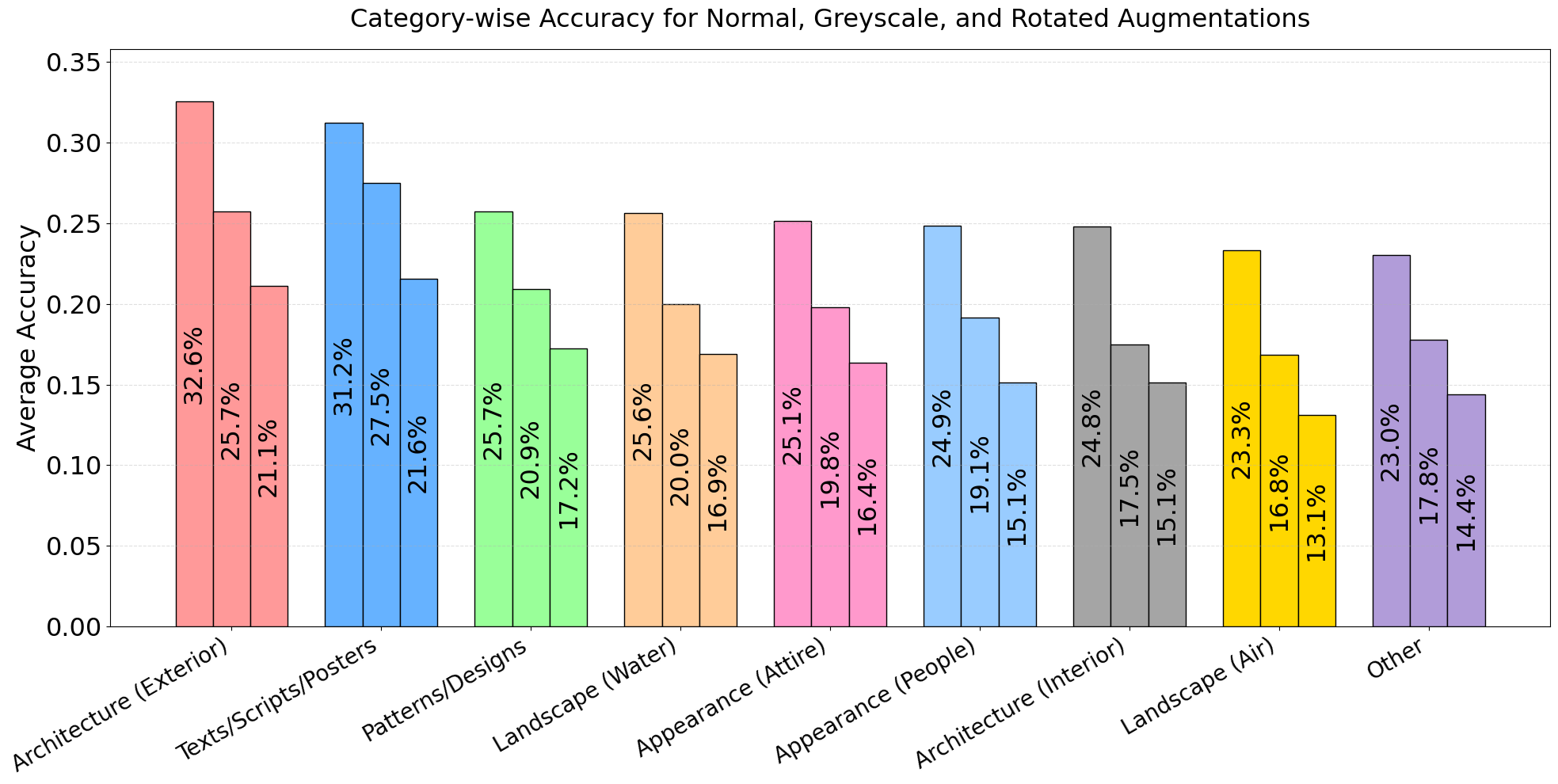}
    \caption{Model-wise averaged accuracy across the nine image categories, as a function of the image perturbations. There is a clear trend of models performing better with the original images (left), compared to the grayscale images (middle) , or rotated images (right).}
    \label{fig:777}
\end{figure}
%%%%%%%%%%%%%%%%%%%%%%%%%%%%%%%%%%%%%%%%%%%%%%%%%%%%%%%%%%%%%%%%%%%%%%%%%%%%%%%%%%%%%%%%%%%%%%%%%%%%%%%%%%%%
\subsection{Effect of Image Perturbations on Results}
\begin{figure}[t!]
    \centering
    \includegraphics[width=1\linewidth]{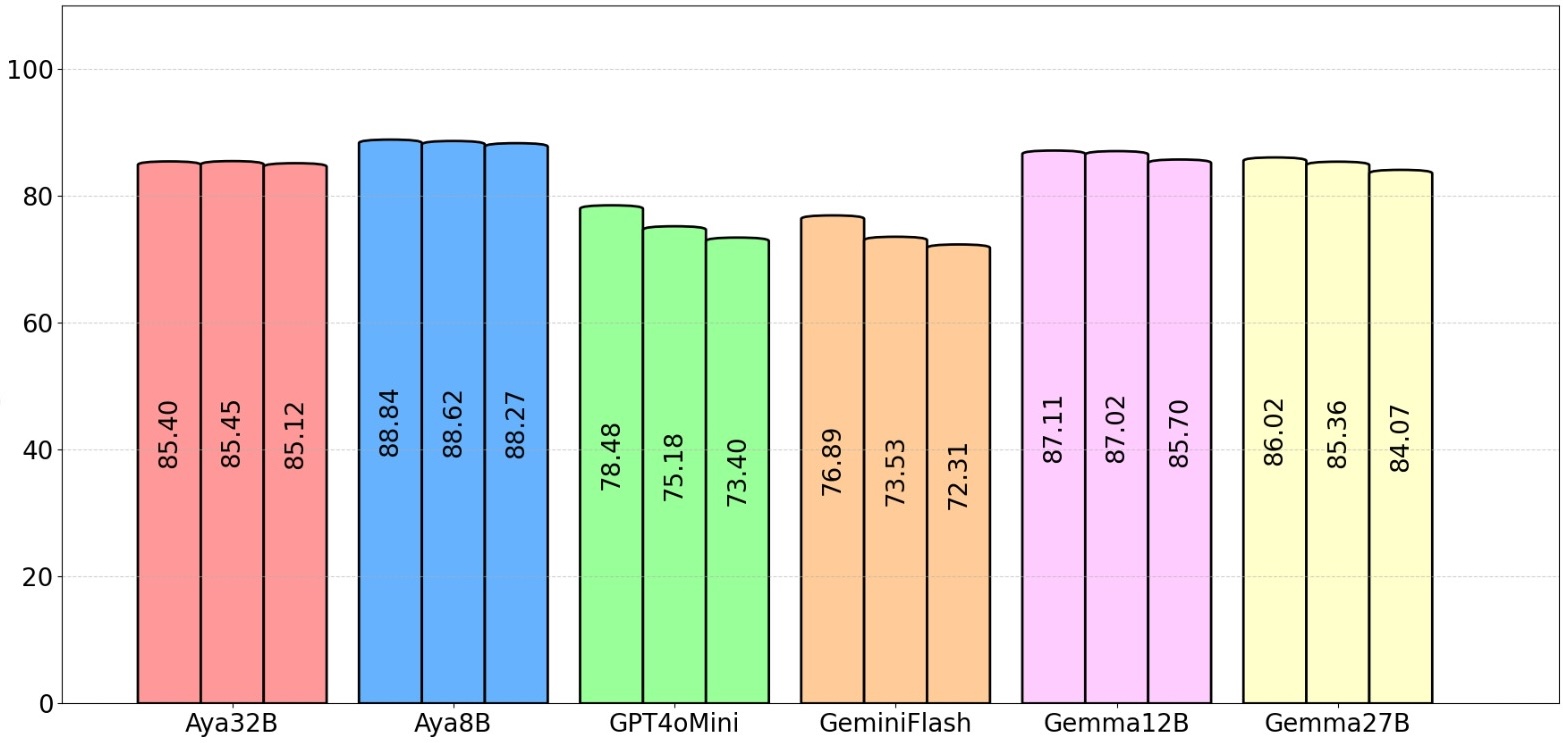}
    \caption{Average model confidence, given the original images (left), grayscale images (middle), and rotated images (right). GPT4o, Gemini-Flash, and Gemma-27B are most sensitive to image perturbations.}
    \label{figure:9}
\end{figure}

\autoref{fig:777} and \autoref{figure:9} display the changes in accuracy observed due to gray-scaling and rotating the images compared to the original images. Input image perturbations can have a large impact on the country-level biases in VLMs. Further, It can be assumed that the VLMs tested are not robust enough towards image perturbations, with each country being effected at a different scale between each model/perturbation. The overall averages can also be seen in \autoref{figure:101}, \autoref{figure:102} and \autoref{figure:103} respectively. 

% The results across all of the VLMs indicate that gray-scaling leads to a minor drop in average accuracy while rotation of images lead to a major drop in average accuracy. But at a country level, only few countries' accuracy were affected considerably by gray-scaling, but image rotation decreases performance uniformly across countries.

~\autoref{figure:4} shows how perturbations affect model performance across different semantic image categories. For all nine categories, models perform best on original (unaltered) images, with decreasing accuracy for gray-scaled and worse for rotated versions. Categories like exterior architecture, text/scripts/posters, and attire/patterns are especially impacted by perturbations. We hypothesize that it is likely because they contain fine-grained, orientation-sensitive, or highly color-dependent details. 

We also look at geographical disparities of these changes in orientation in \autoref{figure:7} and ~\autoref{figure:8}. We observe the disparity in model robustness also emerges clearly. For example, models such as Aya Vision 32B, GPT4o-mini and Gemini 3 12B show very different sensitivity across both a) perturbations and b) regions which were affected. We hypothesise that architectural and training differences might be influencing how models process image orientation and color. While gray-scaling may reduce performance due to the loss of visual detail or color-dependent cues, rotation disrupts spatial reasoning and object orientation, which are critical for geographic or cultural recognition.

These findings highlight the importance of evaluating model performance under realistic image distortions, especially for applications where images may not be clean or consistently formatted as image characteristics can vary widely.

%%%%%%%%%%%%%%%%%%%%%%%%%%%%%%%%%%%%%%%%%%%%%%%%%%%%%%%%%%%%%%%%%%%%%%%%%%%%%%%%%%%%%%%%%%%%%%%%%%%%%%%%%%%%
%%%------------------------------------------------------------------------------------------------------%%%
%%%------------------------------------------------------------------------------------------------------%%%
\begin{figure}
    \centering
    \includegraphics[width=1\linewidth]{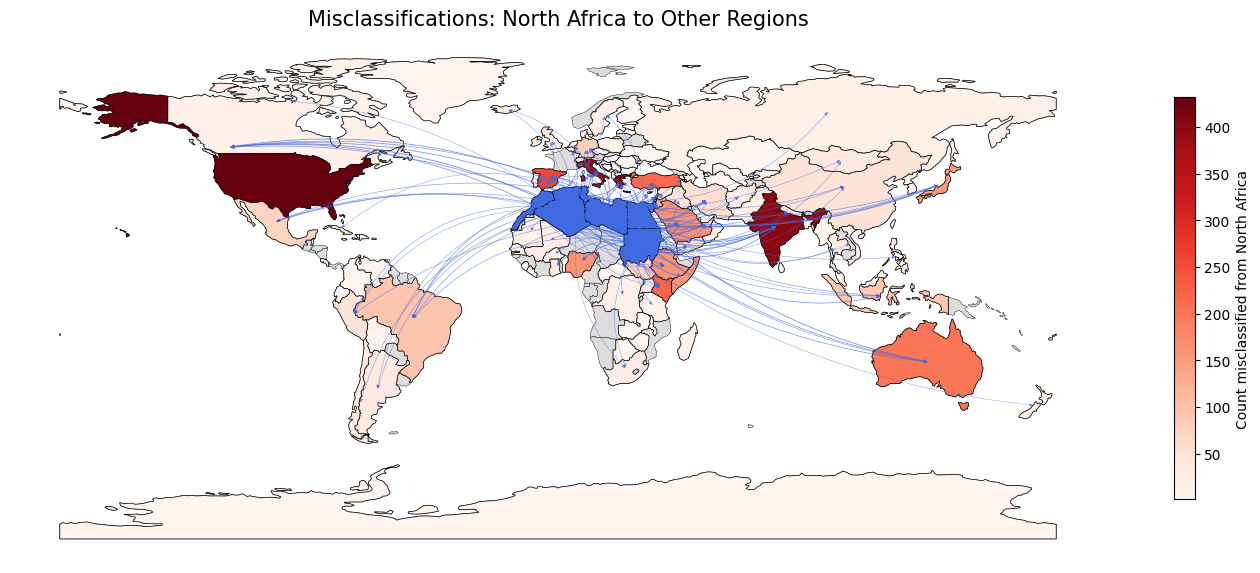}
    \caption{Mis-classification map for North African countries. There is a clear trend of models predicting USA, India, Australia, or geographically close countries in Europe and the Middle East.}
    \label{fig:810}
\end{figure}

\subsection{Effect of Input Variations on Confidence}

%%%------------------------------------------------------------------------------------------------------%%%
\begin{figure*}
    \centering
    \includegraphics[width=0.6\linewidth]{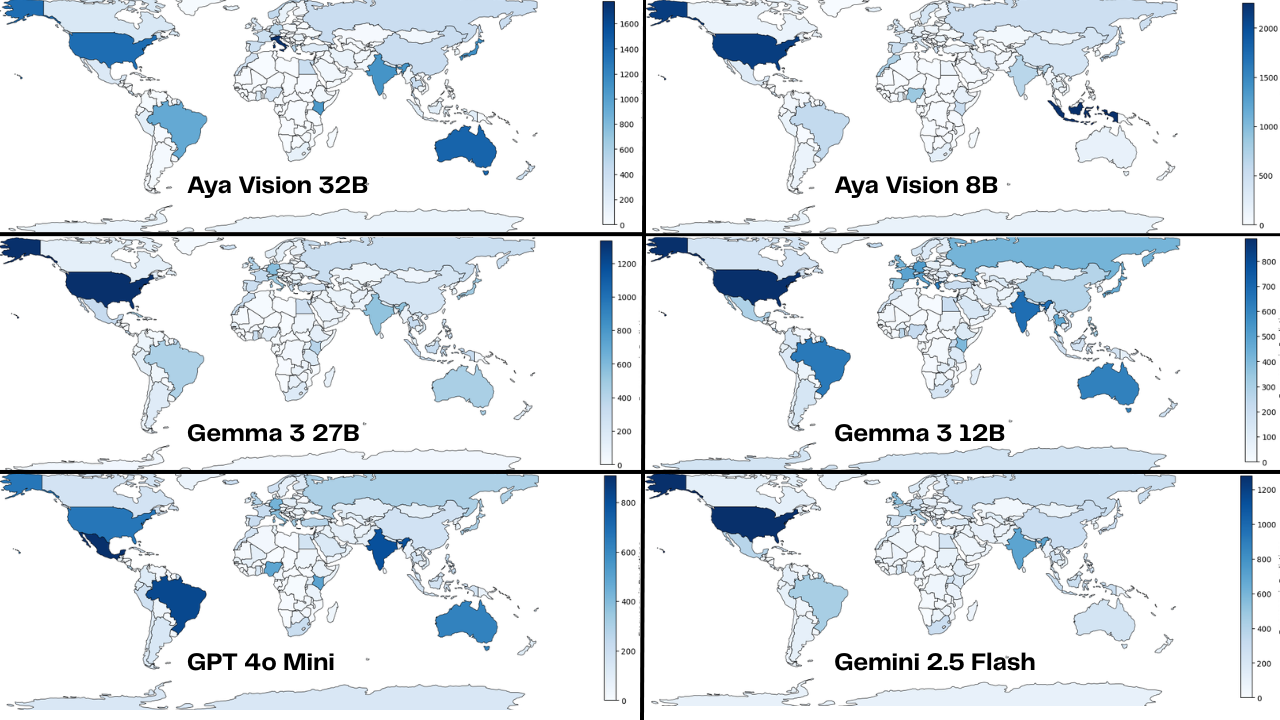}
    \caption{Country-wise response distribution in the open-ended prompt format. There is a consistent trend of models predicting USA, but otherwise, no clear bias towards predicting Western countries.}
    \label{figure:10}
\end{figure*}
Despite the drop in Overall accuracy by all of the tested models due to either of the image perturbations, the confidence of the open-weight models didn't have a significant change while the proprietary models displayed a visible drop in confidence compared to the original images. Compared to rotation of images, Gray-scaling had a larger impact on the response accuracies. The average confidence of each VLM with each adversarial setting compared to the original can be seen in \autoref{figure:9}. The closed-weights models exhibited a drop in confidence when a rotated or grayscale image was provided than the corresponding originals, but this wasn't the case with open-weight models we tested.

%%%%%%%%%%%%%%%%%%%%%%%%%%%%%%%%%%%%%%%%%%%%%%%%%%%%%%%%%%%%%%%%%%%%%%%%%%%%%%%%%%%%%%%%%%%%%%%%%%%%%%%%%%%%
\subsection{Image Feature categories VS accuracy}
Apart from the experiments, the original 21.1 K images were also labeled multi-way based on the key features they contain using larger VLMs like Gemini-2.5-Pro, o4-mini, Grok-2-Vision. Later a majority vote of each label was considered. The quality was later manually verified over a subset by multiple people \footnote{Feature category labels were verified on a subset of 10\% samples equally distributed over all countries, with 2 people verifying labels, in cases  with no consensus between the two, the third annotator was used.}. We have used 9 sub-categories for this categorization. The descriptions of each of these categories can be seen in \autoref{table:3}.
A large variance was observed between each feature category and the country level accuracies obtained. Additionally there was also a large variation between how accuracy was affected for each country/feature based on model/perturbation used. This can be also be seen in \autoref{figure:900}. The extent to which each category's images were recognized by VLMs can be seen in \autoref{fig:777}. External architecture and native language texts' presence in the background helped the VLMs recognize the culture better compared to the other features. 
%\begin{figure*}
%    \centering
%    \includegraphics[width=0.78\linewidth]{AccVScat.png}
%    \caption{Model-wise averaged accuracy across the nine image categories, as a function of the image perturbations. There is a clear trend of models performing better with the original images (left), compared to the grayscale images (middle) , or rotated images (right).}
%    \label{fig:777}
%\end{figure*}
%%%------------------------------------------------------------------------------------------------------%%%
%\begin{figure*}
%    \centering
%    \includegraphics[width=1\linewidth]{BiasEx1.png}
%    \caption{An arrow map displaying what countries' images were classified as India : External Architecture}
%    \label{figure:201}
%\end{figure*}

%%%%%%%%%%%%%%%%%%%%%%%%%%%%%%%%%%%%%%%%%%%%%%%%%%%%%%%%%%%%%%%%%%%%%%%%%%%%%%%%%%%%%%%%%%%%%%%%%%%%%%%%%%%%
\subsection{Distribution of Predicted countries}
The distribution of responses in an open ended approach can be seen in \autoref{figure:10}. The output distributions varied largely among models, even those within the same family (i.e between Gemma-3-27B, Gemma-3-12B and Aya-vision-32B, Aya-vision-8B). The results obtained contradict the usual assumption about western biases in generative models, and was observed over a few nations with likely high training data proportion. 

Notably, all models consistently overpredict certain countries—particularly the USA, India, and Brazil—regardless of actual ground truth. We hypothesize that these countries are likely overrepresented in the models’ pretraining data or benefit from more visually distinctive cues. Biases seem to cluster around a few highly represented or visually salient countries rather than reflecting broader geopolitical landscape.

These results show that model predictions are likely highly influenced by data availability and image characteristics rather than a generic global bias. It also underscores the need for better interpretability regarding the geographic composition of VLM training datasets to fully understand such biases.

%%%%%%%%%%%%%%%%%%%%%%%%%%%%%%%%%%%%%%%%%%%%%%%%%%%%%%%%%%%%%%%%%%%%%%%%%%%%%%%%%%%%%%%%%%%%%%%%%%%%%%%%%%%%
\subsection{Misclassification Analysis}
The mapping of misclassification of samples was not limited to similar or neighboring nations. This can be observed in \autoref{fig:801} to \autoref{fig:816}. These misclassifications varied by each individual feature and provide a better fine-grained insights of cultural biases. For instance, Apart from neighboring / similar countries, most images from Africa and rural regions of South America were classified as India. A specific example is shown in \autoref{fig:810} where out of the 600 images (100 * 6 models), roughly 80-120 belong to this category for most countries, while many countries had most of their misclassified as originating from India.
%%%%%%%%%%%%%%%%%%%%%%%%%%%%%%%%%%%%%%%%%%%%%%%%%%%%%%%%%%%%%%%%%%%%%%%%%%%%%%%%%%%%%%%%%%%%%%%%%%%%%%%%%%%%
%DE: we should not reference figures in the Appendix in the conclusion

\section{Discussion}

%%%------------------------------------------------------------------------------------------------------%%%
Our study presents a comprehensive analysis of cultural biases in Vision-Language Models (VLMs) using a geographically balanced dataset across 211 countries. We evaluated popular models across multiple prompting strategies, e.g. open-ended, multiple-choice (random and similar distractors), and multilingual settings. Open-ended formats showed the greatest disparity in country-level accuracy, particularly in underrepresented regions such as Central Africa and parts of South America.The use of culturally similar distractors proved to be the most effective in revealing fine-grained errors, highlighting limitations in models' cultural discrimination abilities.

We further assessed the models' robustness to image perturbations like gray-scaling and rotation. While gray-scaling affected only a few specific countries, rotation led to a broad and uniform drop in performance, confirming that VLMs rely heavily on image orientation. We further observed that performance also varied by semantic image content—categories like architecture, textual cues, and attire were more predictive of cultural origin, especially in unaltered images. Language variation in prompts had minimal impact on average accuracy, though countries closely tied to the input language (e.g., Spain with Spanish, Brazil with Portuguese) showed slight gains. However, this trend was inconsistent and did not generalize across all culturally linked regions.

Finally, our misclassification analysis shows that models frequently confuse images from low-resource or visually ambiguous countries with a few dominant nations, reinforcing the role of training data bias. 

\section{Conclusion}
Our findings show that biases are not uniformly Western but instead reflect over representation of certain countries in training data. Model performance varied across prompt types, languages, image features, and perturbations—highlighting limitations in robustness and cultural generalization. These results call for greater transparency in dataset composition and the need for more culturally inclusive evaluation methods to ensure fairer and more globally representative VLMs.

\section*{Limitations}
Our study has a few important limitations to keep in mind. First, the use of country-level labels as a proxy for culture, while common for large-scale analysis, inherently overlooks intra-country cultural diversity and multicultural populations, potentially obscuring sub-national or regional nuances. The country labels used don’t account for political complexities like disputed territories.
%%%%%%%%%%%%%%%%%%%%%%%%%%%%%%%%%%%%%%%%%%%%%%%%%%%%%%%%%%%%%%%%%%%%%%%%%%%%%%%%%%%%%%%%%%%%%%%%%%%%%%%%%%%%
%\section*{Ethics Statement}
%Cultural Biases are a sensitive topic, and hence potential overgeneralisations and cultural biases could lead to undesirable outcomes. 
%%%%%%%%%%%%%%%%%%%%%%%%%%%%%%%%%%%%%%%%%%%%%%%%%%%%%%%%%%%%%%%%%%%%%%%%%%%%%%%%%%%%%%%%%%%%%%%%%%%%%%%%%%%%
\section*{Acknowledgements}
This work was partially supported by a research grant from Cohere Labs.
%%%%%%%%%%%%%%%%%%%%%%%%%%%%%%%%%%%%%%%%%%%%%%%%%%%%%%%%%%%%%%%%%%%%%%%%%%%%%%%%%%%%%%%%%%%%%%%%%%%%%%%%%%%%
\bibliography{anthology,custom}
\bibliographystyle{acl_natbib}
%%%%%%%%%%%%%%%%%%%%%%%%%%%%%%%%%%%%%%%%%%%%%%%%%%%%%%%%%%%%%%%%%%%%%%%%%%%%%%%%%%%%%%%%%%%%%%%%%%%%%%%%%%%%
\appendix
%%%%%%%%%%%%%%%%%%%%%%%%%%%%%%%%%%%%%%%%%%%%%%%%%%%%%%%%%%%%%%%%%%%%%%%%%%%%%%%%%%%%%%%%%%%%%%%%%%%%%%%%%%%%
%%%%%%%%%%%%%%%%%%%%%%%%%%%%%%%%%%%%%%%%%%%%%%%%%%%%%%%%%%%%%%%%%%%%%%%%%%%%%%%%%%%%%%%%%%%%%%%%%%%%%%%%%%%%
%%%%%%%%%%%%%%%%%%%%%%%%%%%%%%%%%%%%%%%%%%%%%%%%%%%%%%%%%%%%%%%%%%%%%%%%%%%%%%%%%%%%%%%%%%%%%%%%%%%%%%%%%%%%
\section{Overall Accuracies Before and After Image Perturbations}
\label{sec:appendix-A}
\begin{figure*}
    \centering
    \includegraphics[width=0.82\linewidth]{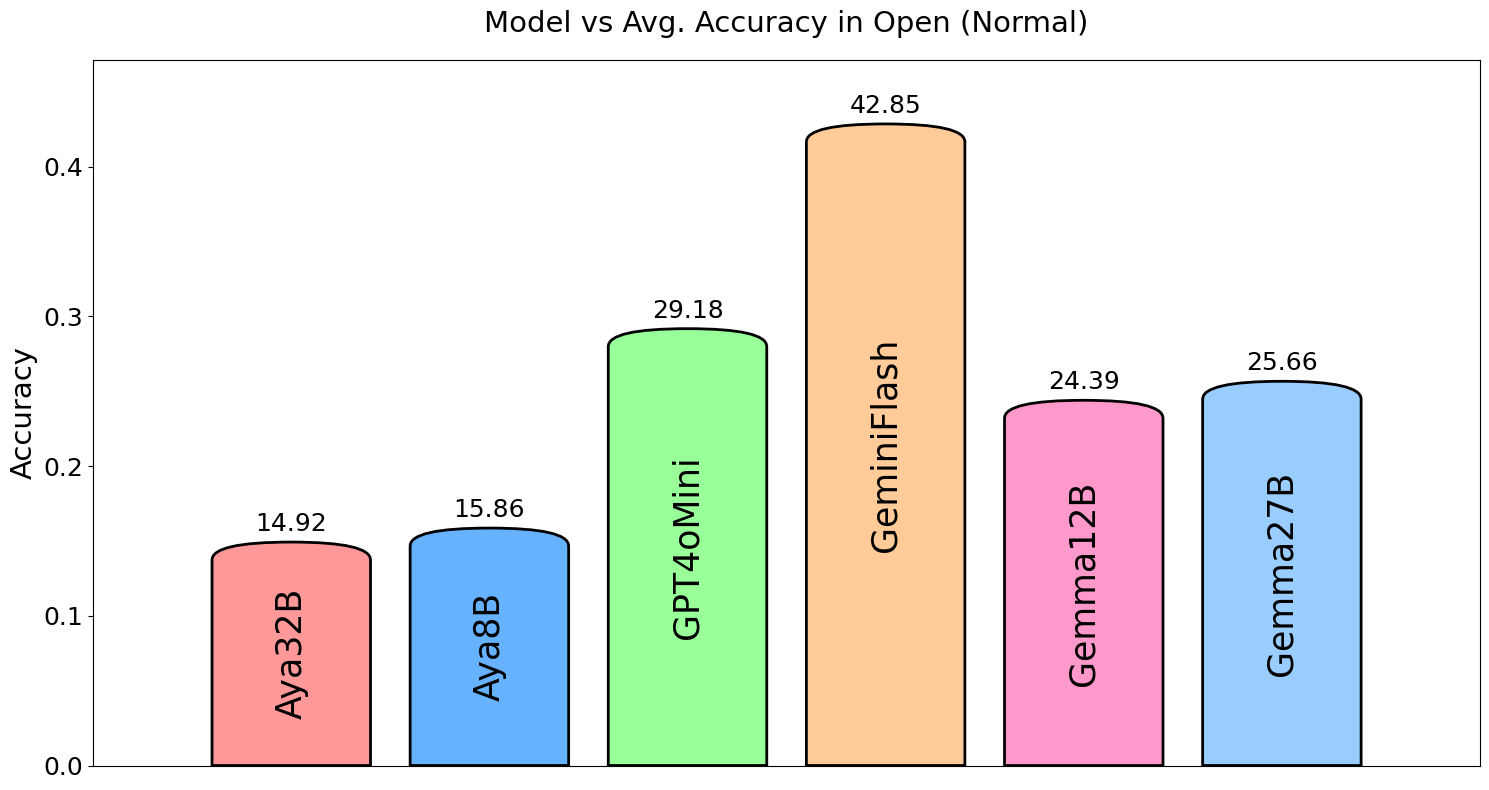}
    \caption{Overall Accuracy : Open Ended (Normal)}
    \label{figure:101}
\end{figure*}

\begin{figure*}
    \centering
    \includegraphics[width=0.82\linewidth]{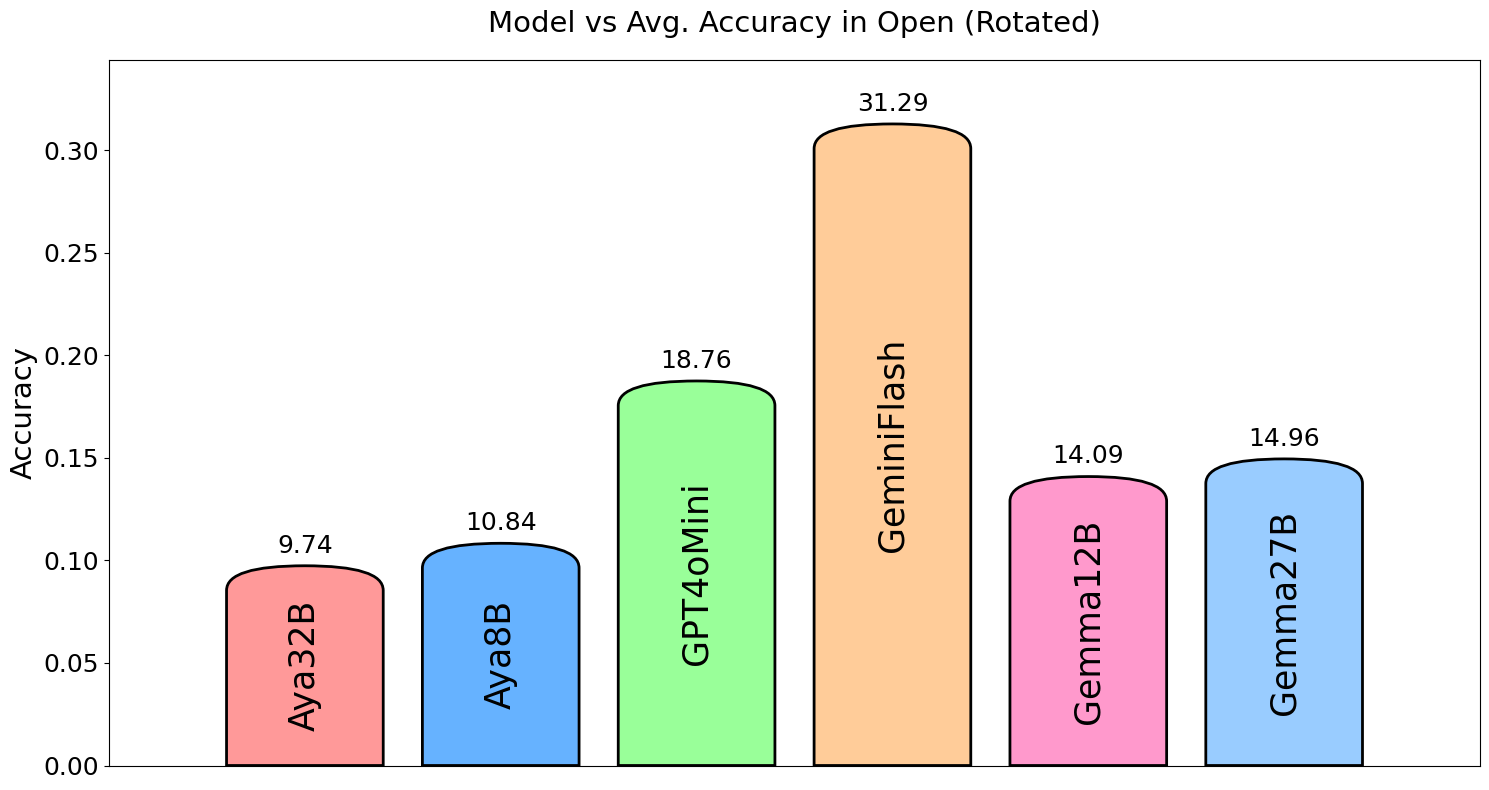}
    \caption{Overall Accuracy : Open Ended (Rotated)}
    \label{figure:102}
\end{figure*}

\begin{figure*}
    \centering
    \includegraphics[width=0.82\linewidth]{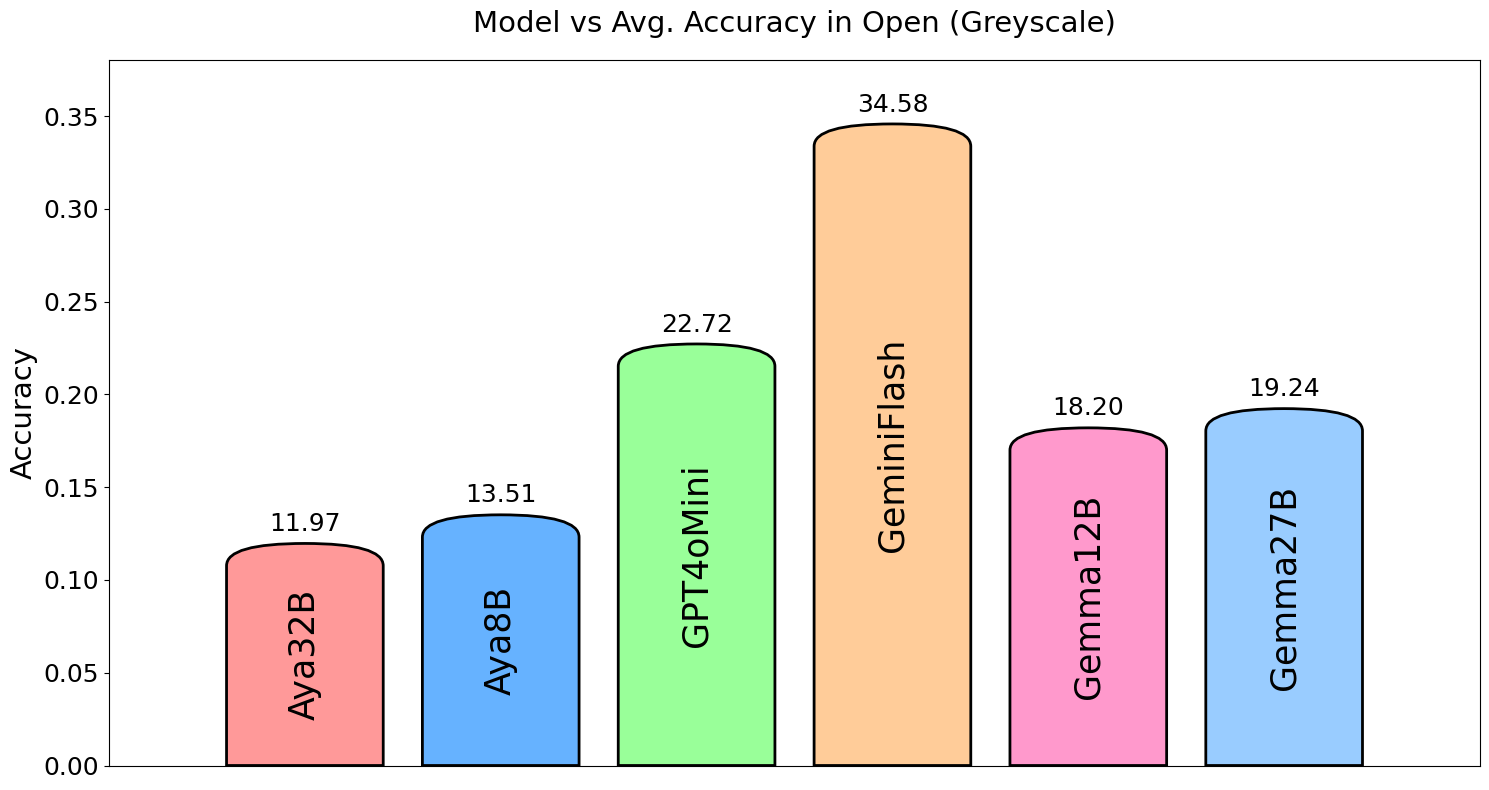}
    \caption{Overall Accuracy : Open Ended (Grayscale)}
    \label{figure:103}
\end{figure*}

\autoref{figure:101}, \autoref{figure:102}, \autoref{figure:103} display the accuracy obtained for each image perturbation used compared to the original through each of the VLMs tested.
%%%%%%%%%%%%%%%%%%%%%%%%%%%%%%%%%%%%%%%%%%%%%%%%%%%%%%%%%%%%%%%%%%%%%%%%%%%%%%%%%%%%%%%%%%%%%%%%%%%%%%%%%%%%
\section{Overall Accuracy VS Models used : In each MCQ setting}
\begin{figure}
    \centering
    \includegraphics[width=0.82\linewidth]{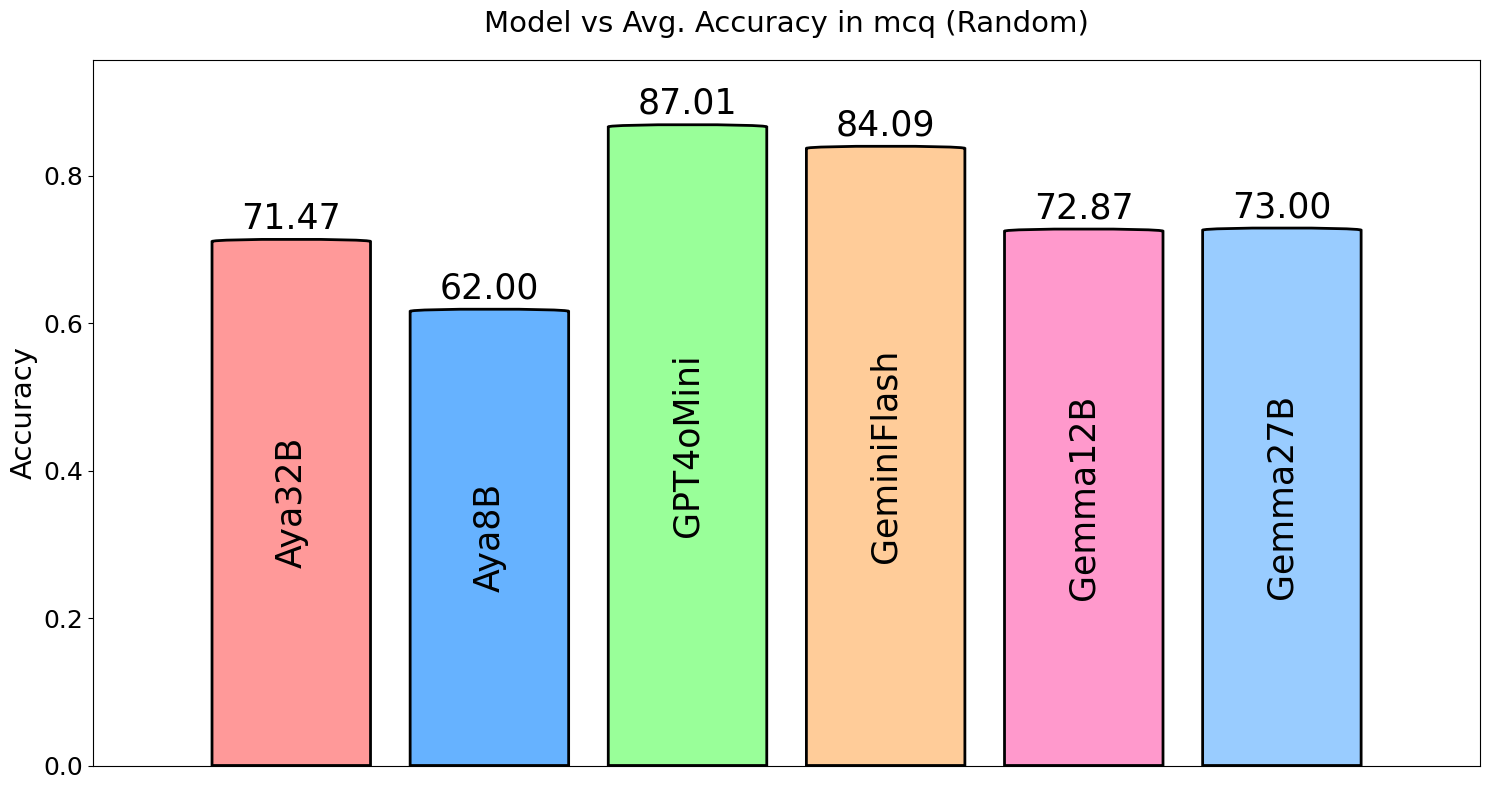}
    \caption{Overall Accuracy : MCQ-Random : Model wise}
    \label{figure:104}
\end{figure}

\begin{figure}
    \centering
    \includegraphics[width=0.82\linewidth]{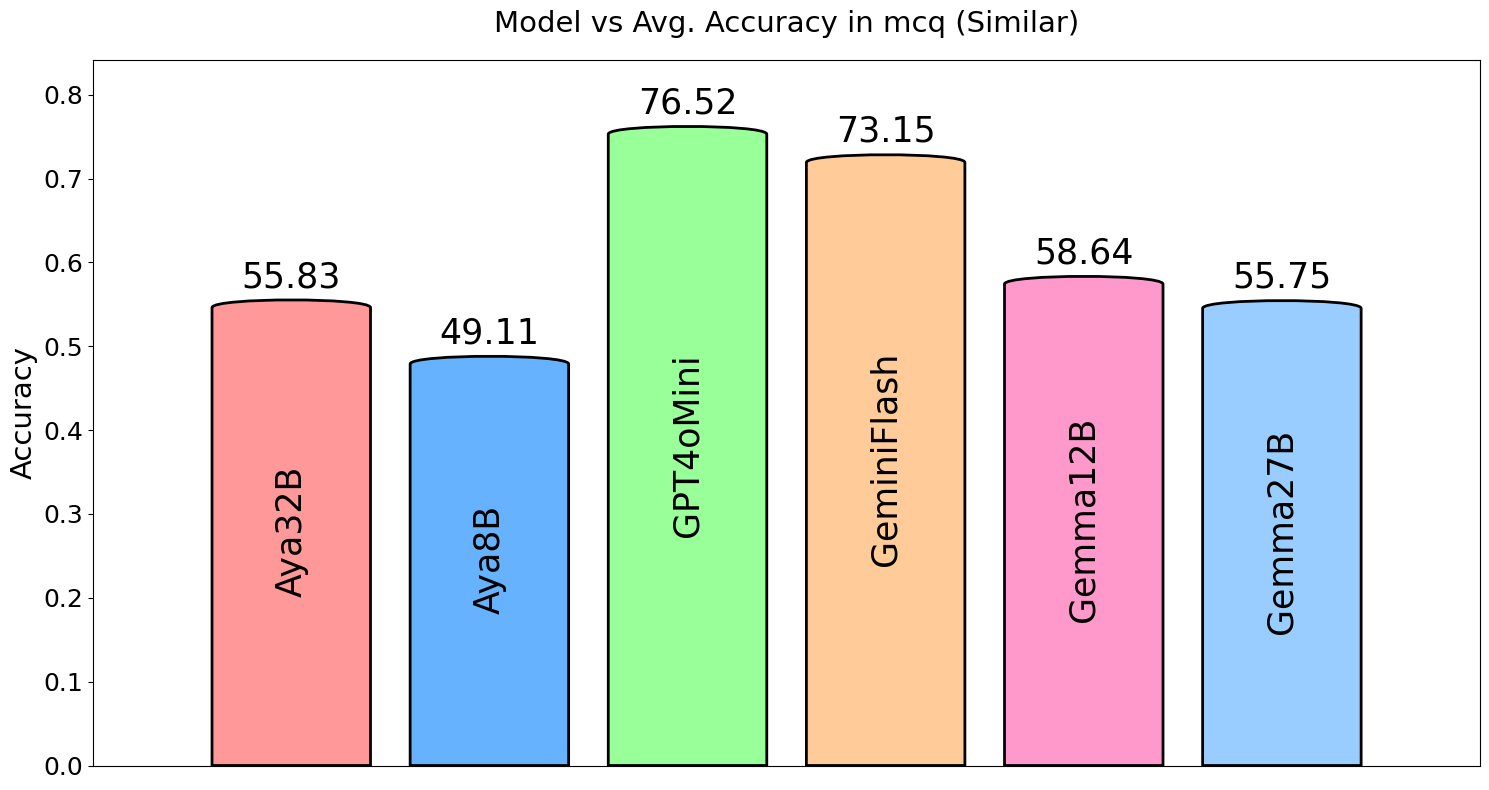}
    \caption{Overall Accuracy : MCQ-Similar : Model wise}
    \label{figure:105}
\end{figure}

\autoref{figure:104}, \autoref{figure:105} display the accuracy obtained through each model in each MCQ experiment.
\label{sec:appendix-B}
%%%%%%%%%%%%%%%%%%%%%%%%%%%%%%%%%%%%%%%%%%%%%%%%%%%%%%%%%%%%%%%%%%%%%%%%%%%%%%%%%%%%%%%%%%%%%%%%%%%%%%%%%%%%
\section{Reproducibility}
\label{sec:appendix-C}
Inference was done through Cohere's API for Aya-Vision-8B and Aya-Vision-32B through the default hyperparameters with a seed value of 1024. The rest of the models were used through OpenRouter's API through the default hyper-parameters with a seed value of 1024. The experiments were repeated thrice and the overall accuracy varied between 1-1.2\%, with some countries' accuracy varying up to 1.5\%. The costs associated with all experiments combined were 850\$ through OpenRouter and 250\$ Cohere API credits. The experiments were run on TPUs costing 0.35\$/hr with the costs reaching ~60\$. 
%%%%%%%%%%%%%%%%%%%%%%%%%%%%%%%%%%%%%%%%%%%%%%%%%%%%%%%%%%%%%%%%%%%%%%%%%%%%%%%%%%%%%%%%%%%%%%%%%%%%%%%%%%%%
\begin{table*}
    \centering
    \begin{tabular}{lp{12cm}}
    \hline
    \rowcolor{purple2} \textbf{\small{Category}} & \textbf{\small{Description}} \\
    \hline
    \small{Appearance (Attire)} & \small{Attires of some people from the image, clothes being hanged in the background, etc.} \\
    \small{Appearance (People)} & \small{Appearance / visual perception of people's ethnicity, presence of any celebrities, etc.} \\
    \small{Architecture (Exterior)} & \small{Building facades, monuments, bridges, outdoor structures, and any external architectural elements visible in the scene.} \\
    \small{Architecture (Interior)} & \small{Indoor environments e.g. rooms, corridors, staircases, furniture, and interior design details.} \\
    \small{Landscape (Water)} & \small{Bodies of water such as oceans, rivers, lakes, waterfalls, ponds, and any aquatic scenery.} \\
    \small{Landscape (Air)} & \small{Aerial / bird’s-eye views, landscapes captured from above, clouds, sky scenes, and horizon vistas.} \\
    \small{Landscape (Vegetation)} & \small{Forests, grasslands, gardens, crops, shrubs, foliage patterns, plant life, or visible greenery.} \\
    \small{Texts/Scripts/Posters} & \small{Signs, banners, billboards, labels, handwritten or printed text, posters, and any other written or graphic messaging.} \\
    \small{Patterns/Designs} & \small{Decorative motifs, surface textures, fabric prints, wallpaper or tile patterns, abstract designs, and repetitive graphical elements.} \\
    \hline
    \end{tabular}
    \caption{Overview of the image categories used to analyse model performance as a function of the type of image.}
    \label{table:3}
\end{table*}

%%%%%%%%%%%%%%%%%%%%%%%%%%%%%%%%%%%%%%%%%%%%%%%%%%%%%%%%%%%%%%%%%%%%%%%%%%%%%%%%%%%%%%%%%%%%%%%%%%%%%%%%%%%%

\section{Prompt Usage}
\label{sec:appendix-D}
\lstset{
    basicstyle=\small\ttfamily,  % Smaller font and monospaced
    breaklines=true,             % Automatically break lines
    breakatwhitespace=true,      % Only break at whitespace
    frame=single,                % Optional: adds a frame around the prompt
    columns=fullflexible         % Fix spacing issues
}
The prompts used for each experiment setting can be seen below.
\begin{lstlisting}
    ## OPEN ENDED
    Respond in the following format :
    {
      "Country": {
      "type": "string",
      "description": "Just Country Name, Nothing else."
      },
      "Reasoning": {
      "type": "string",
      "description": "Why you feel the image is from that particular country. Explain what features/objects/items/visuals of the image made you think so. Should be fine-grained and clear, but in just 2-3 sentences."
      },
      "Confidence": {
      "type": "integer",
      "description": "A integer score on a scale of 0 to 100 on how confident you are about the classification."
      },
      "Features": {
      "type": "string",
      "description": "Which of the features of the image were primarily used for determining the country labels, separated by commas, ordered with primary feature in the beginning."
      }
    }
\end{lstlisting}
%%%------------------------------------------------------------------------------------------------------%%%
\begin{lstlisting}
    ## MCQ
    prompt_header="Respond with a label from the choices : What country is this image likely from ?\n"
    for l,c in zip(labels,choices):prompt_header+=f"{l}) : {c}\n"
    fixed_prompt=prompt_header+"""Respond in the following format :
    {
      "Label": "Uppercase Alphabet",
      "Country": "The chosen label's country name exactly as it was",
      "Confidence": "Integer between 0 to 100 in numeric format"
    }"""
\end{lstlisting}
%%%%%%%%%%%%%%%%%%%%%%%%%%%%%%%%%%%%%%%%%%%%%%%%%%%%%%%%%%%%%%%%%%%%%%%%%%%%%%%%%%%%%%%%%%%%%%%%%%%%%%%%%%%%
\section{Other Plots}
\label{sec:appendix-J}
%%%------------------------------------------------------------------------------------------------------%%%
\begin{figure*}
    \centering
    \includegraphics[width=1\linewidth]{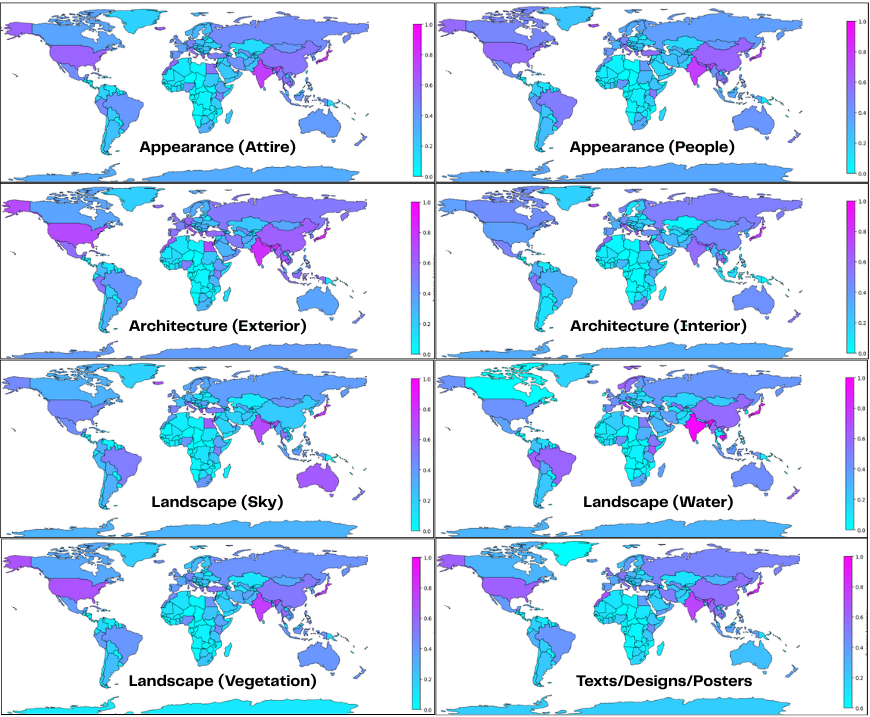}
    \caption{Image Feature categories VS Country wise Accuracy}
    \label{figure:900}
\end{figure*}
%%%------------------------------------------------------------------------------------------------------%%%
\begin{figure*}
    \centering
    \includegraphics[width=0.86\linewidth]{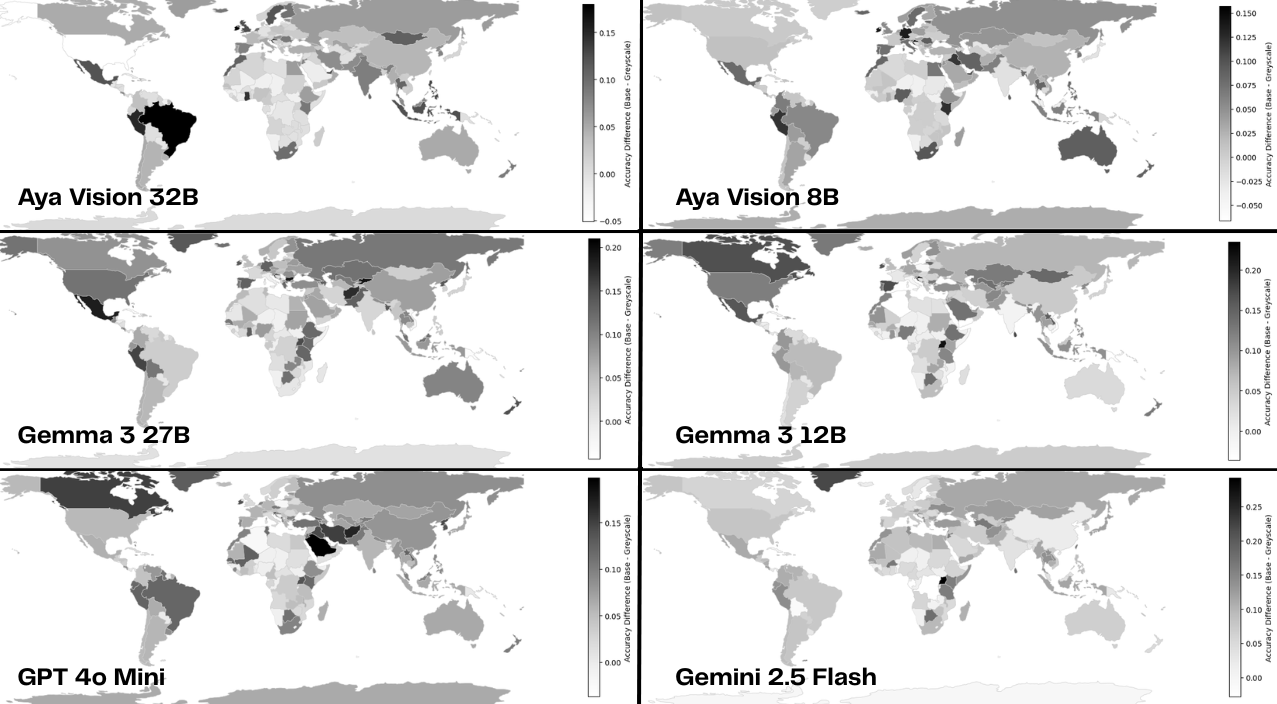}
    \caption{Effect of Gray-scaling VS change in country wise accuracies}
    \caption*{Higher Contrast = Larger Drop in accuracy}
    \label{figure:7}
\end{figure*}
%%%------------------------------------------------------------------------------------------------------%%%
\begin{figure*}
    \centering
    \includegraphics[width=0.86\linewidth]{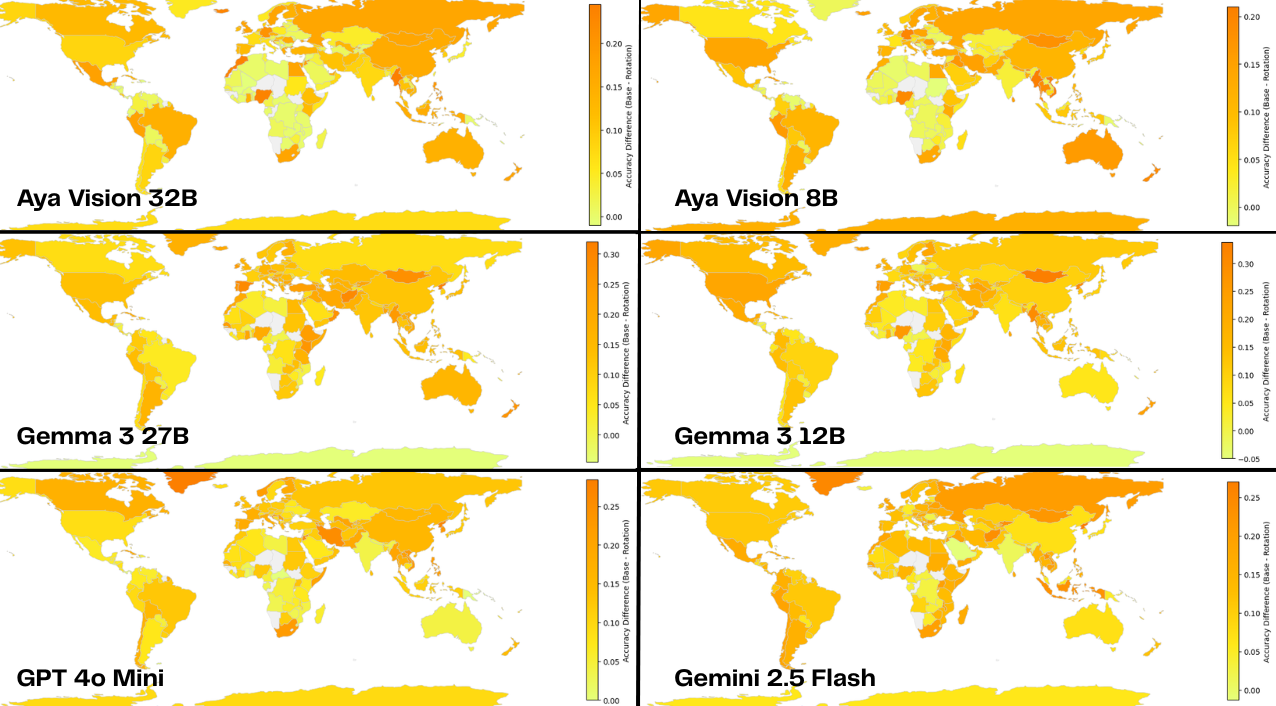}
    \caption{Effect of Rotation VS change in country wise accuracies}
    \caption*{Higher Contrast = Larger Drop in accuracy}
    \label{figure:8}
\end{figure*}
\clearpage
%%%------------------------------------------------------------------------------------------------------%%%
\begin{figure*}
    \centering
    \includegraphics[width=1\linewidth]{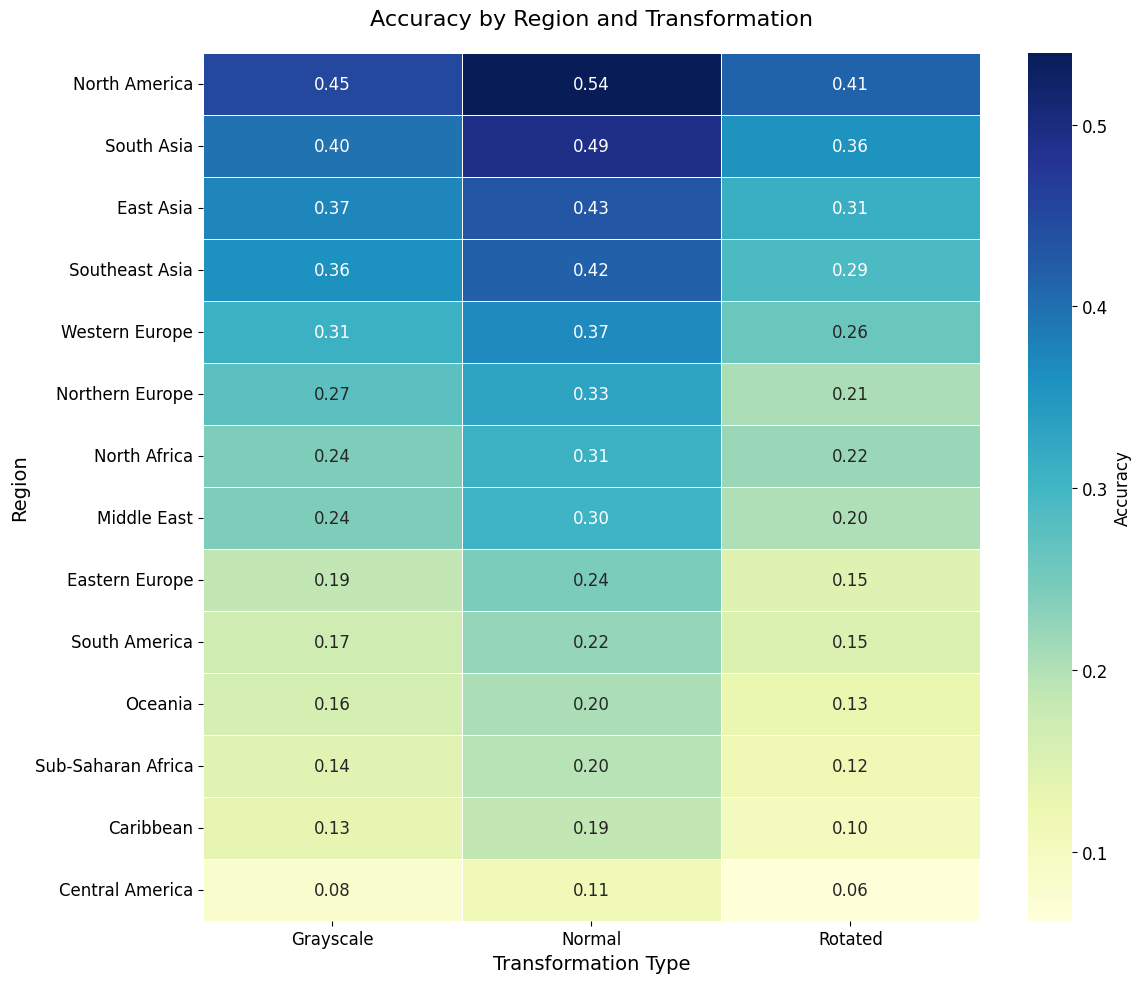}
    \caption{Region wise effect of perturbations}
    \label{figure:666}
\end{figure*}
%%%------------------------------------------------------------------------------------------------------%%%
\begin{figure*}
    \centering
    \includegraphics[width=0.82\linewidth]{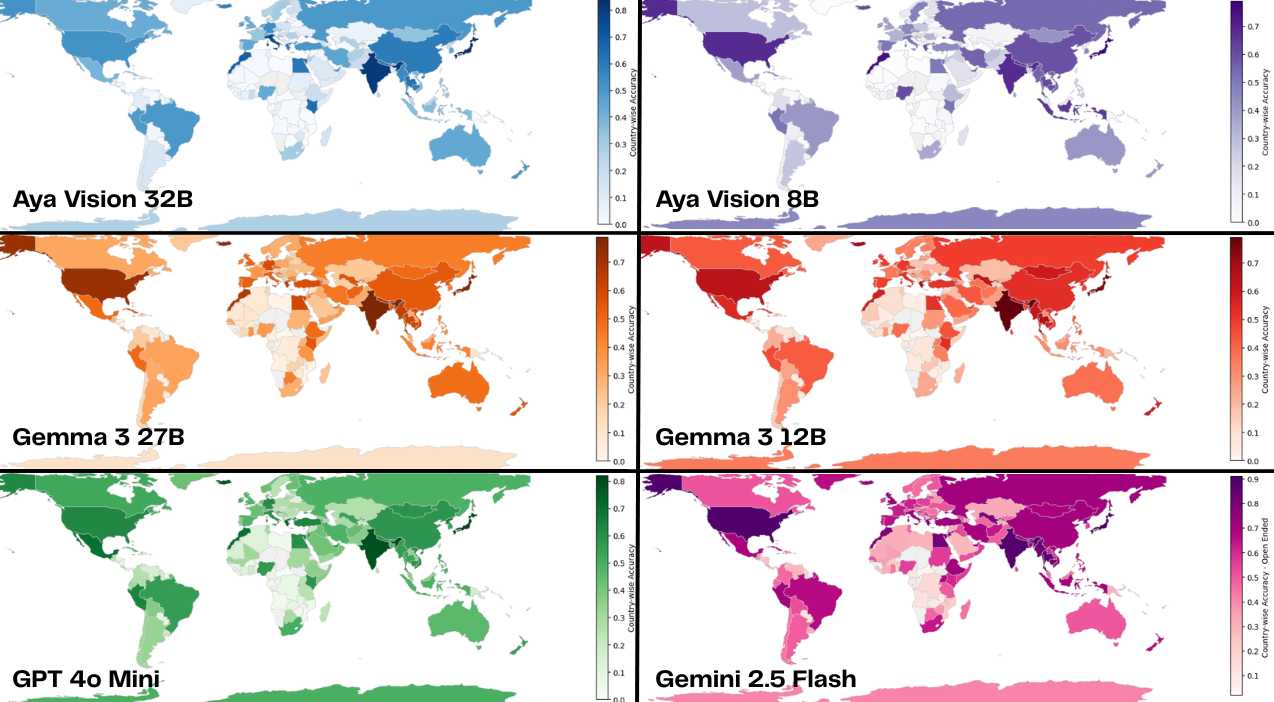}
    \caption{Accuracy over each country's images through open-ended Experiments}
    \label{figure:3}
\end{figure*}
%%%------------------------------------------------------------------------------------------------------%%%
\begin{figure*}
    \centering
    \includegraphics[width=0.82\linewidth]{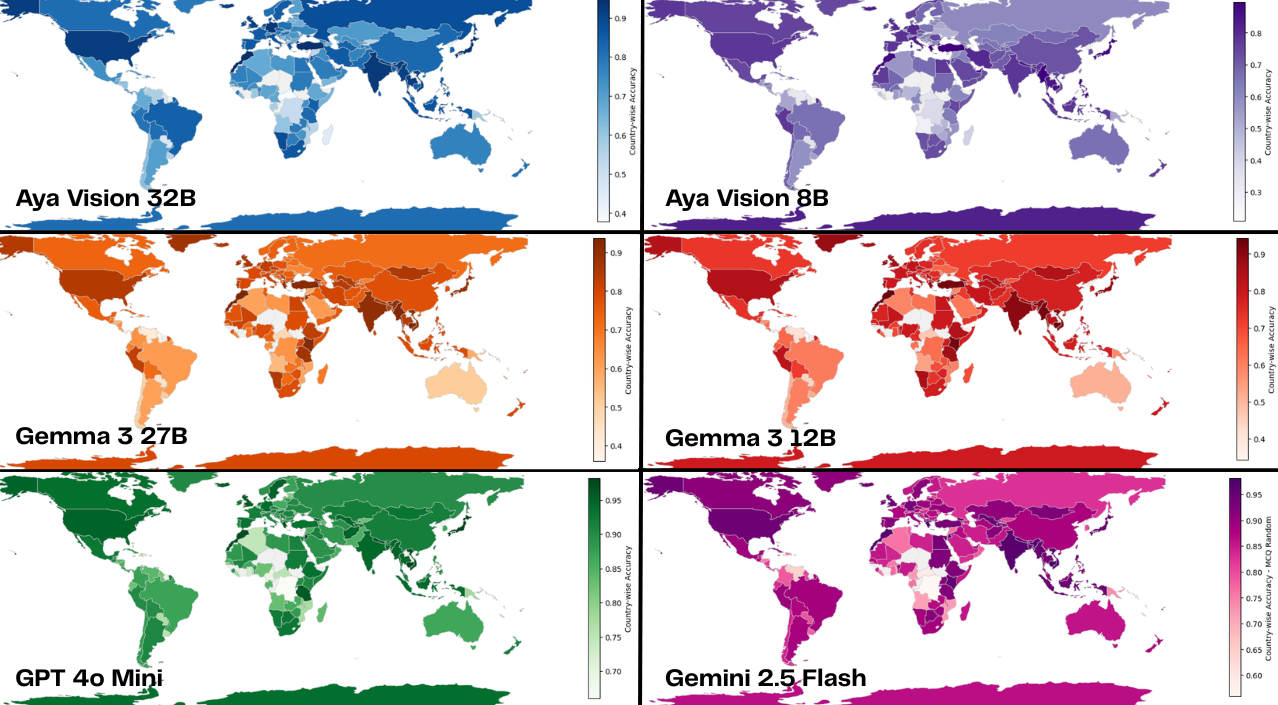}
    \caption{Accuracy over each country's images through MCQ Experiments with random distractors}
    \label{figure:4}
\end{figure*}
%%%------------------------------------------------------------------------------------------------------%%%
\begin{figure*}
    \centering
    \includegraphics[width=0.82\linewidth]{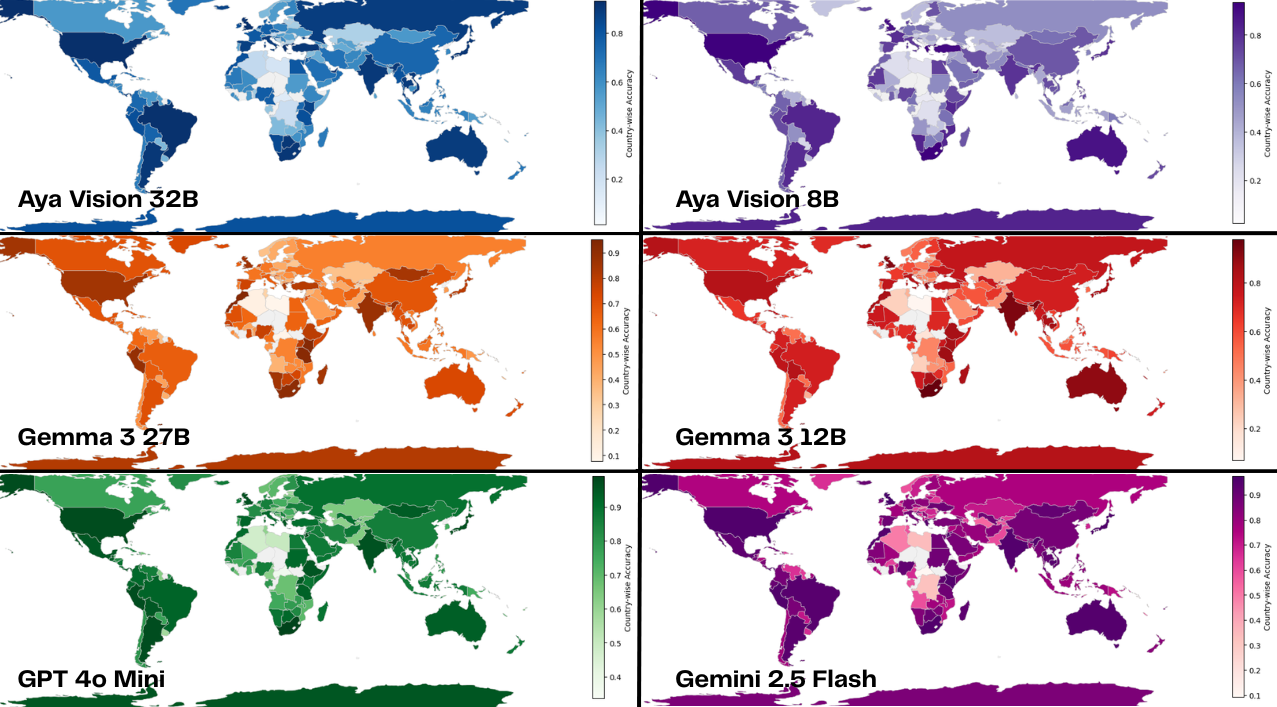}
    \caption{Accuracy over each country's images through MCQ Experiments with similar distractors}
    \label{figure:5}
\end{figure*}
%%%%%%%%%%%%%%%%%%%%%%%%%%%%%%%%%%%%%%%%%%%%%%%%%%%%%%%%%%%%%%%%%%%%%%%%%%%%%%%%%%%%%%%%%%%%%%%%%%%%%%%%%%%%
\section{Mis-Classification Map : Region-wise}
\label{sec:appendix-I}
\begin{figure*}
    \centering
    \includegraphics[width=1\linewidth]{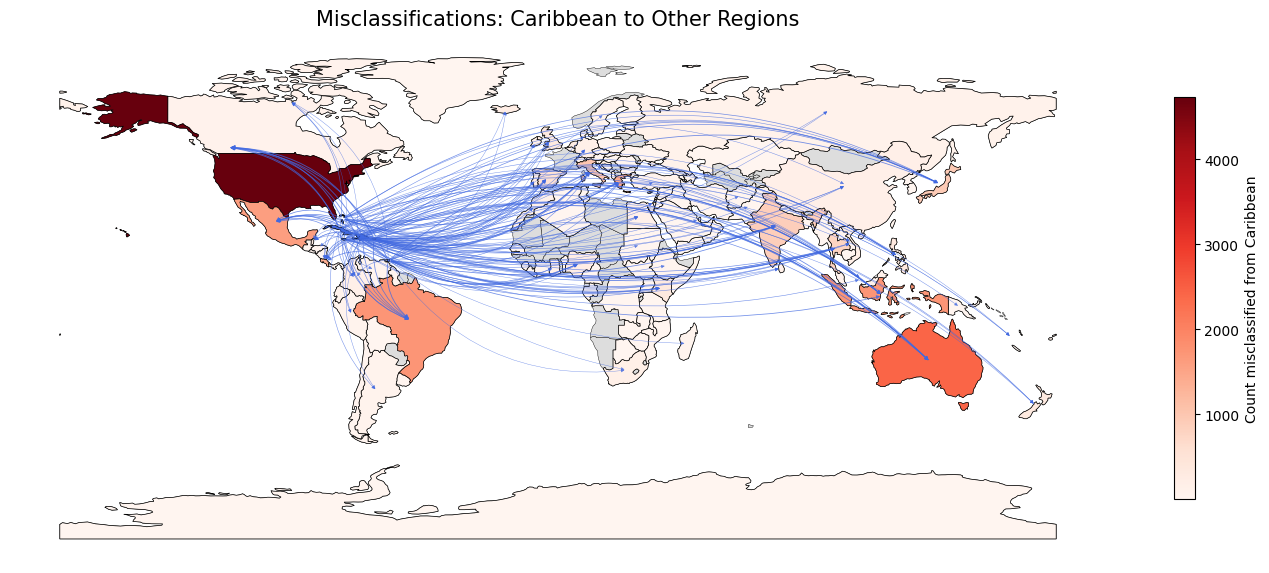}
    \caption{Mis-classification map : Caribbean}
    \label{fig:801}
\end{figure*}

\begin{figure*}
    \centering
    \includegraphics[width=1\linewidth]{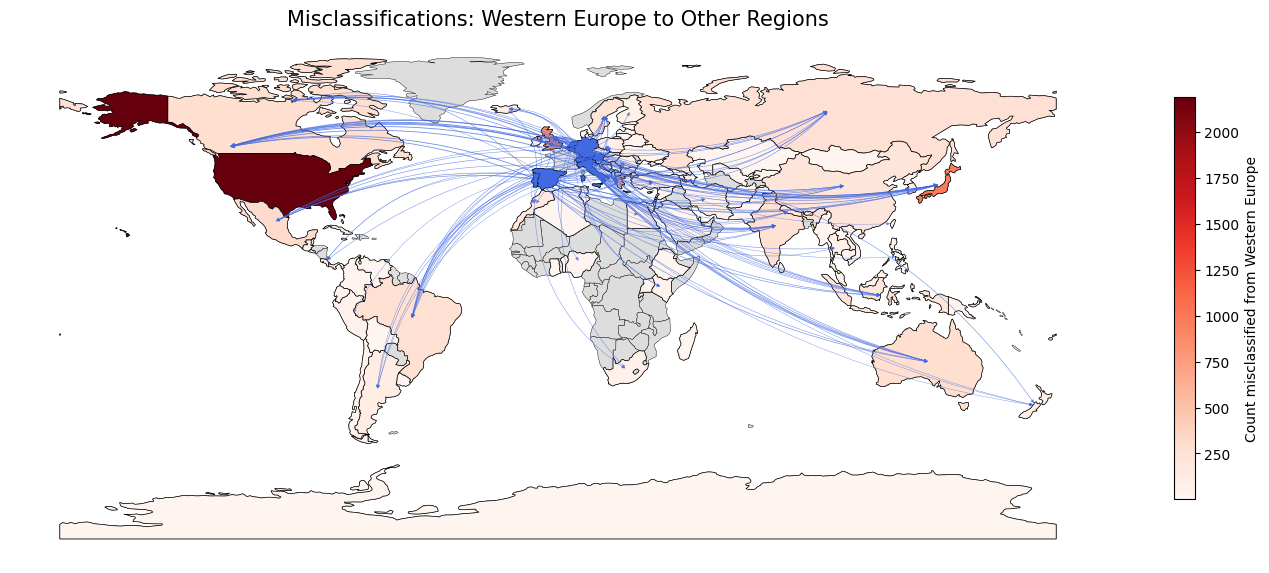}
    \caption{Mis-classification map : Western Europe}
    \label{fig:802}
\end{figure*}

\begin{figure*}
    \centering
    \includegraphics[width=1\linewidth]{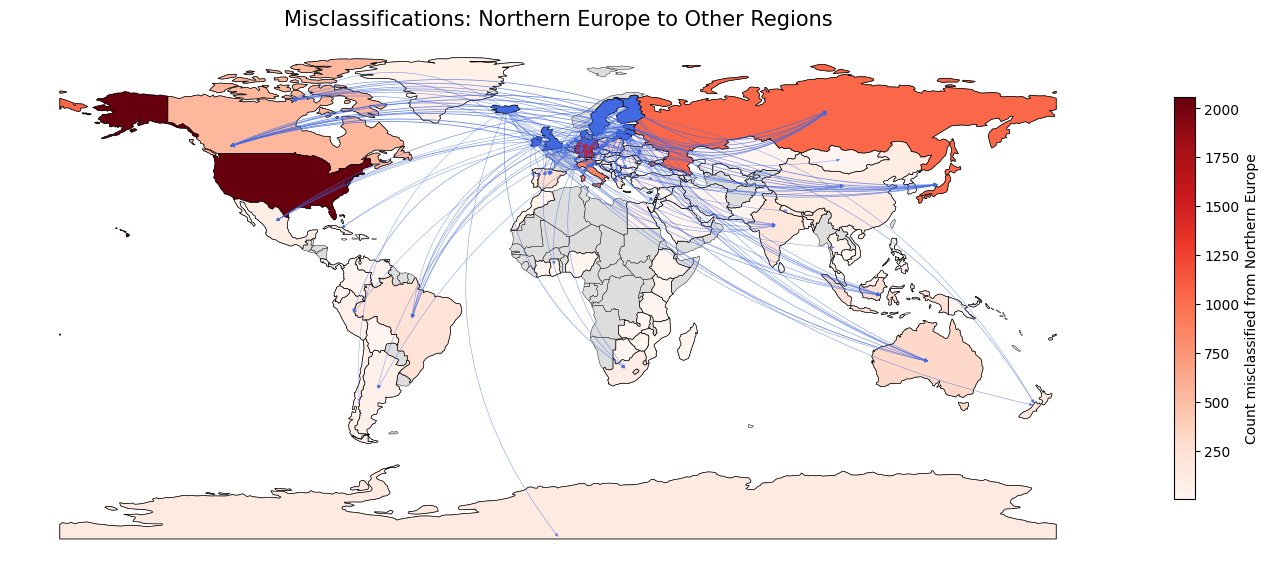}
    \caption{Mis-classification map : North Europe}
    \label{fig:803}
\end{figure*}

\begin{figure*}
    \centering
    \includegraphics[width=1\linewidth]{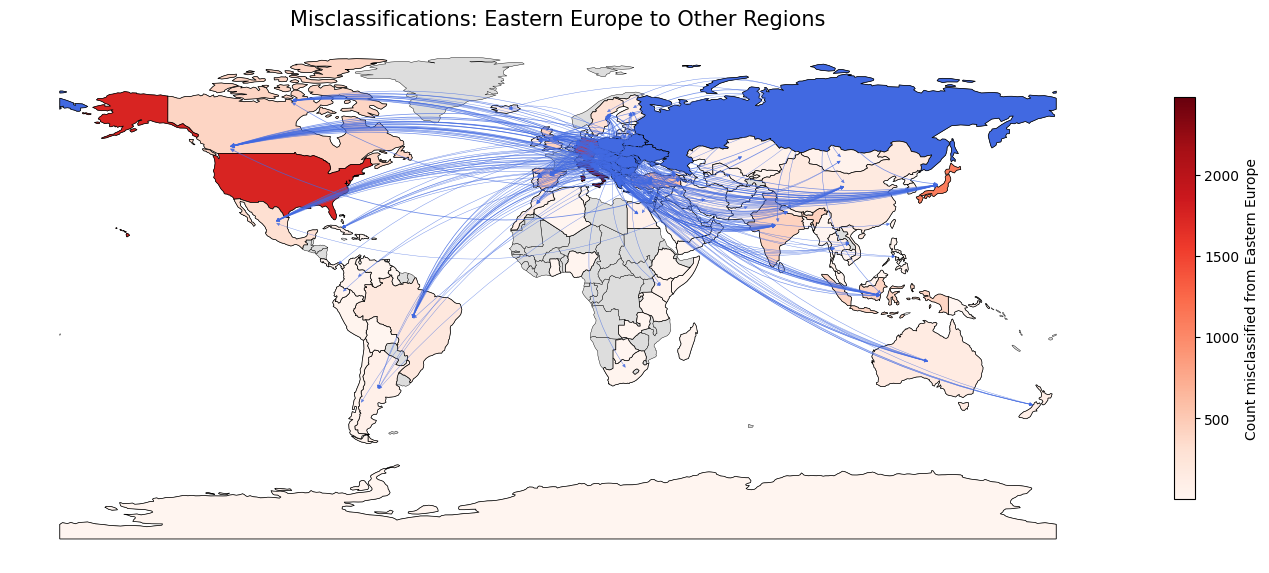}
    \caption{Mis-classification map : Eastern Europe}
    \label{fig:804}
\end{figure*}

\begin{figure*}
    \centering
    \includegraphics[width=1\linewidth]{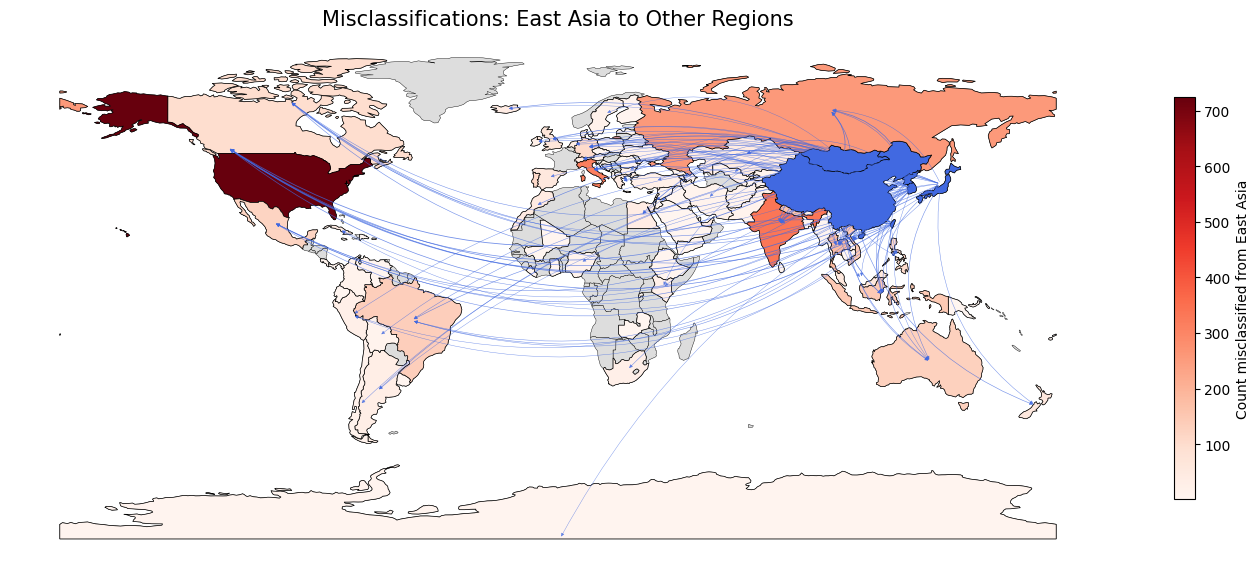}
    \caption{Mis-classification map : East Asia}
    \label{fig:805}
\end{figure*}

\begin{figure*}
    \centering
    \includegraphics[width=1\linewidth]{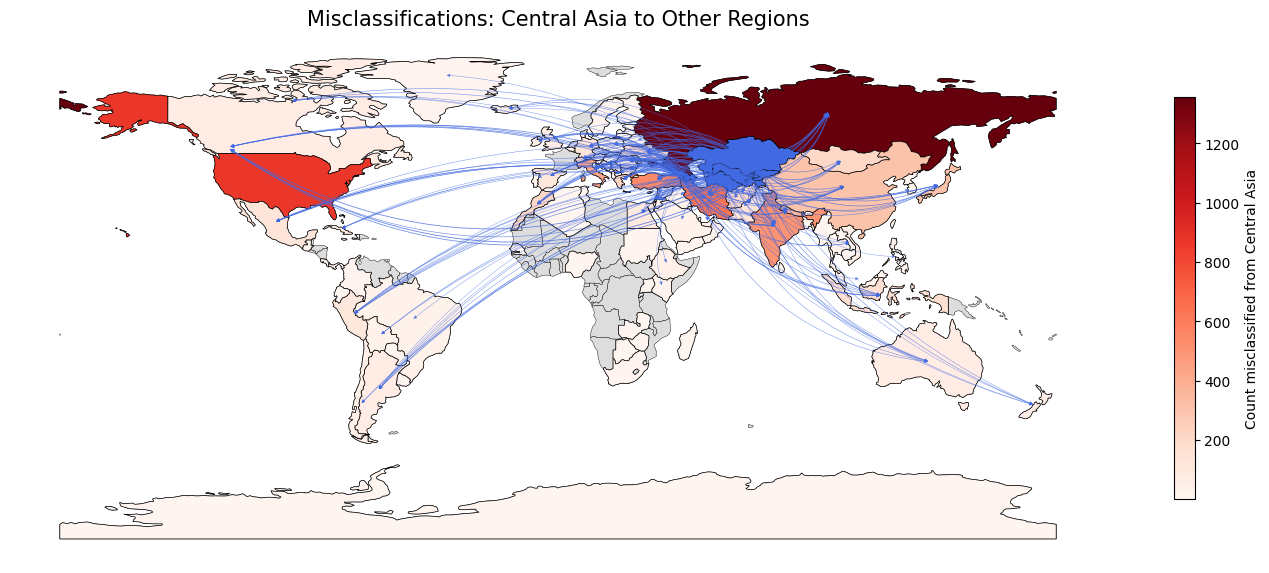}
    \caption{Mis-classification map : Central Asia}
    \label{fig:806}
\end{figure*}

\begin{figure*}
    \centering
    \includegraphics[width=1\linewidth]{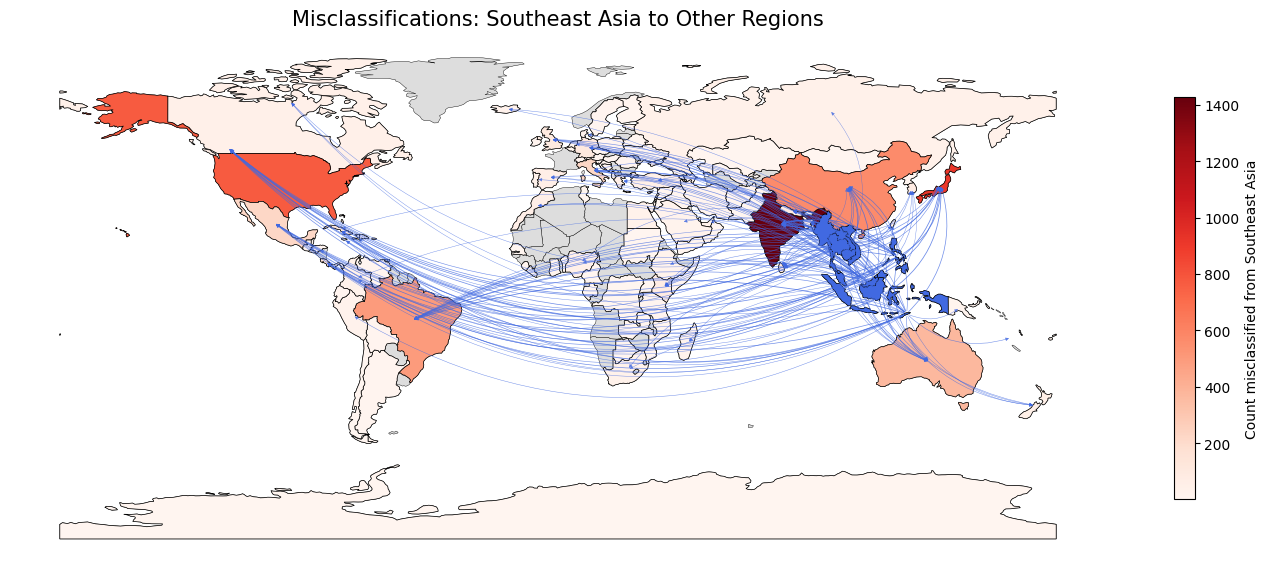}
    \caption{Mis-classification map : South East Asia}
    \label{fig:807}
\end{figure*}

\begin{figure*}
    \centering
    \includegraphics[width=1\linewidth]{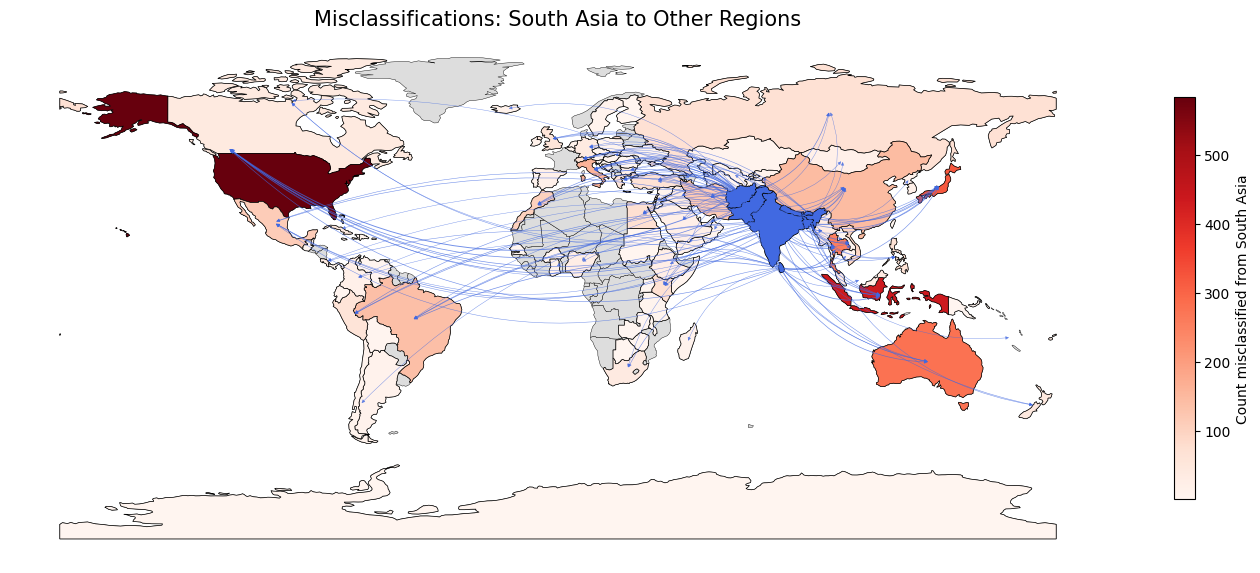}
    \caption{Mis-classification map : South Asia}
    \label{fig:808}
\end{figure*}

\begin{figure*}
    \centering
    \includegraphics[width=1\linewidth]{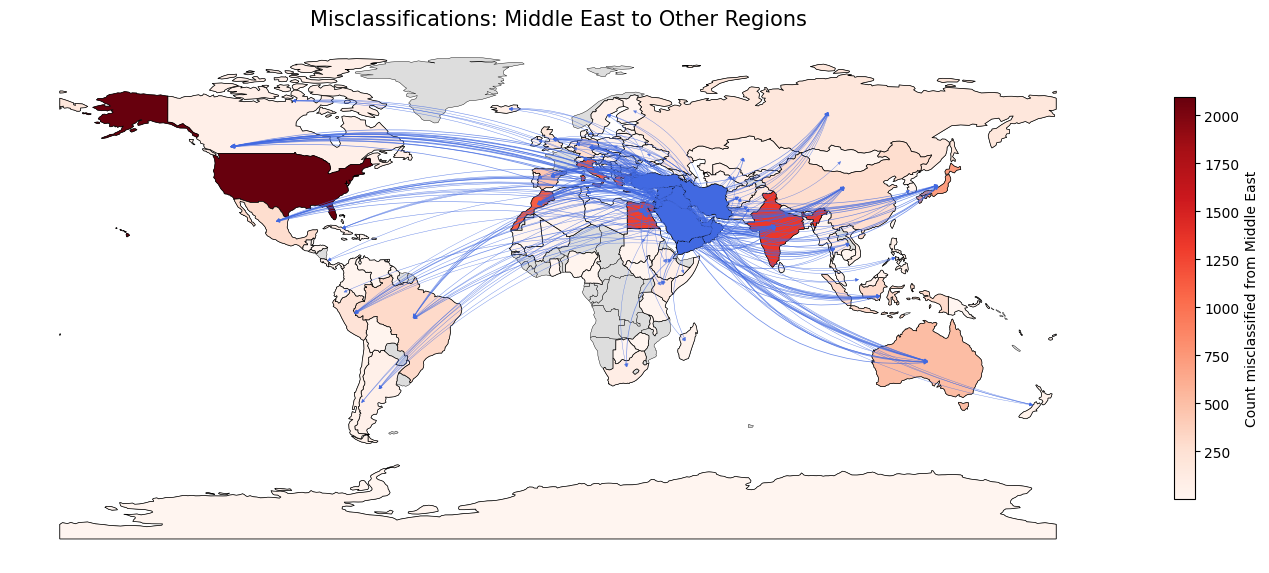}
    \caption{Mis-classification map : Middle East}
    \label{fig:809}
\end{figure*}

\begin{figure*}
    \centering
    \includegraphics[width=1\linewidth]{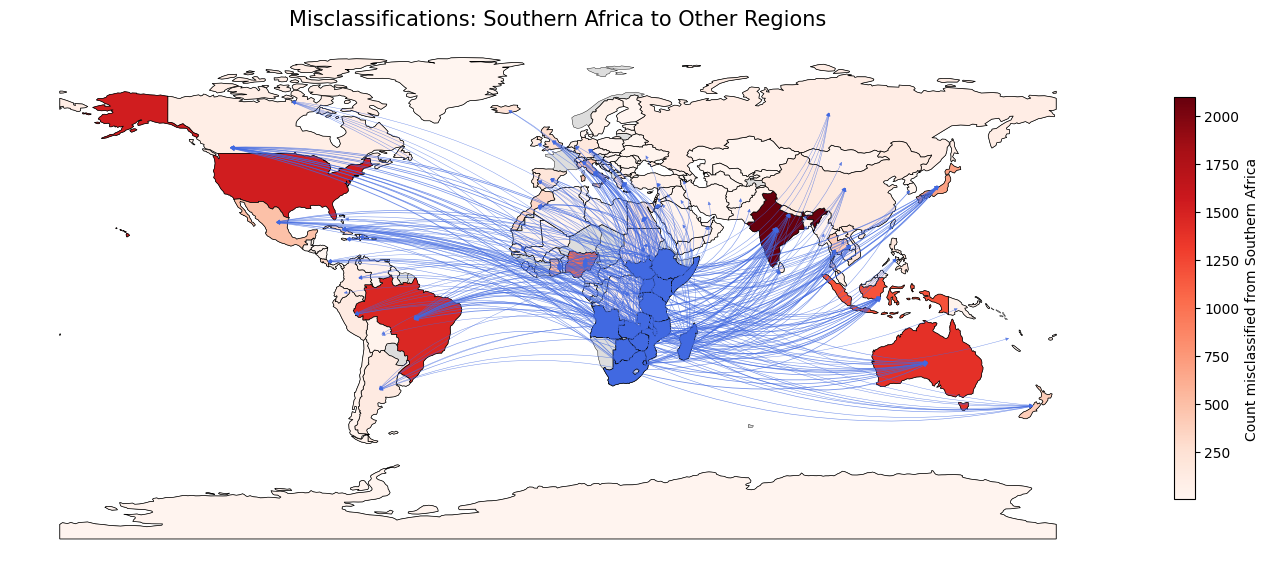}
    \caption{Mis-classification map : Southern Africa}
    \label{fig:811}
\end{figure*}

\begin{figure*}
    \centering
    \includegraphics[width=1\linewidth]{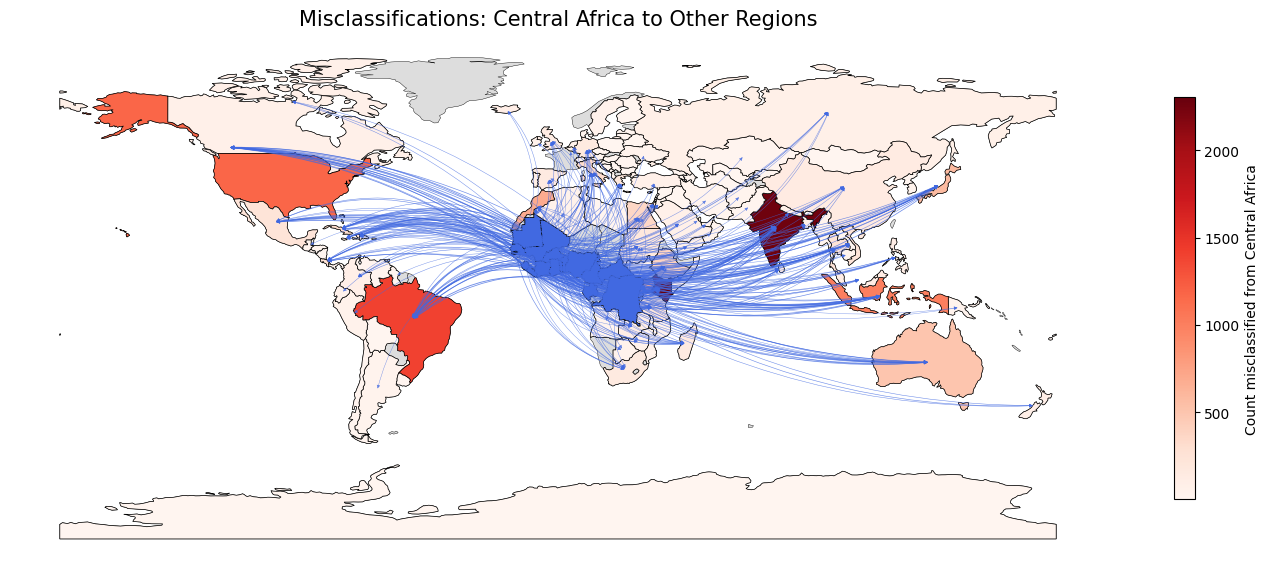}
    \caption{Mis-classification map : Central Africa}
    \label{fig:812}
\end{figure*}

\begin{figure*}
    \centering
    \includegraphics[width=1\linewidth]{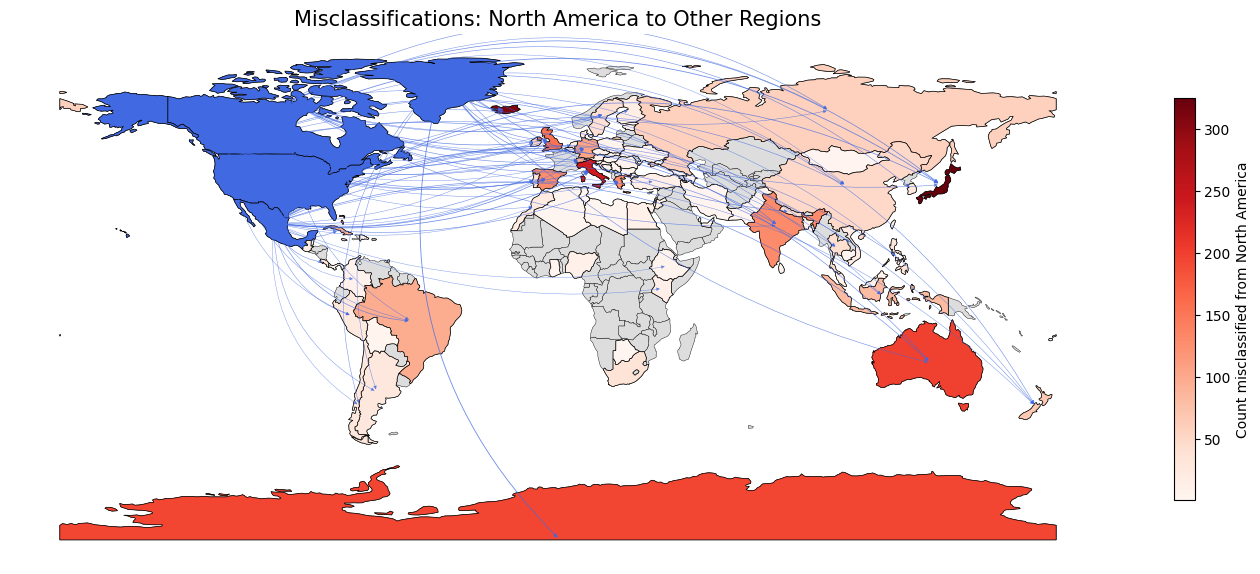}
    \caption{Mis-classification map : North America}
    \label{fig:813}
\end{figure*}

\begin{figure*}
    \centering
    \includegraphics[width=0.95\linewidth]{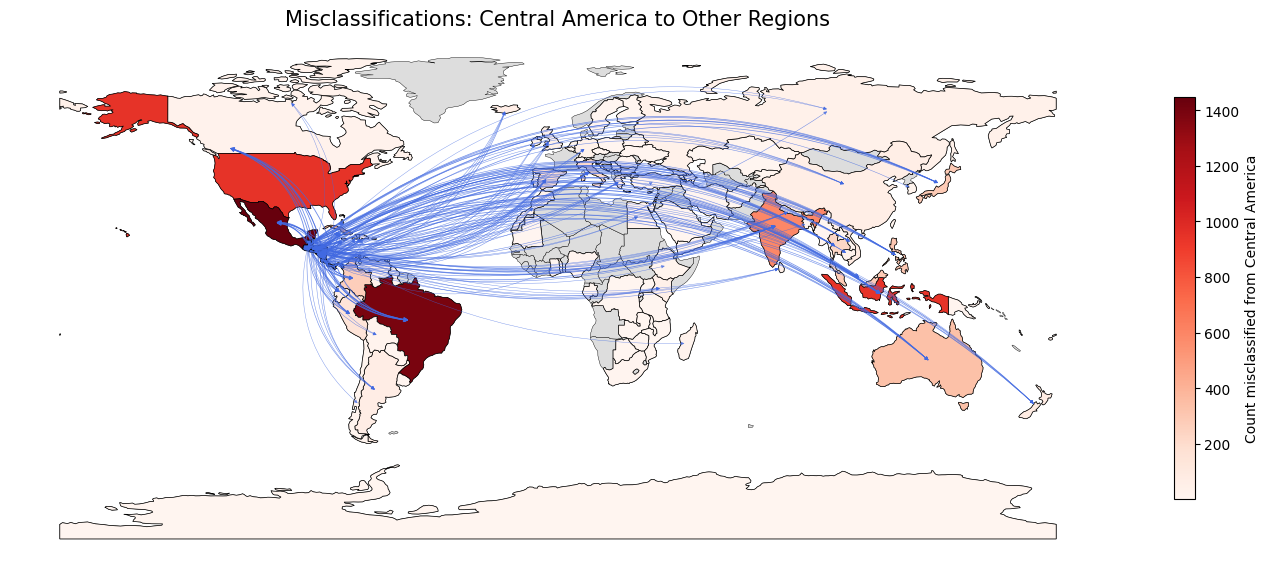}
    \caption{Mis-classification map : Central America}
    \label{fig:814}
\end{figure*}

\begin{figure*}
    \centering
    \includegraphics[width=0.95\linewidth]{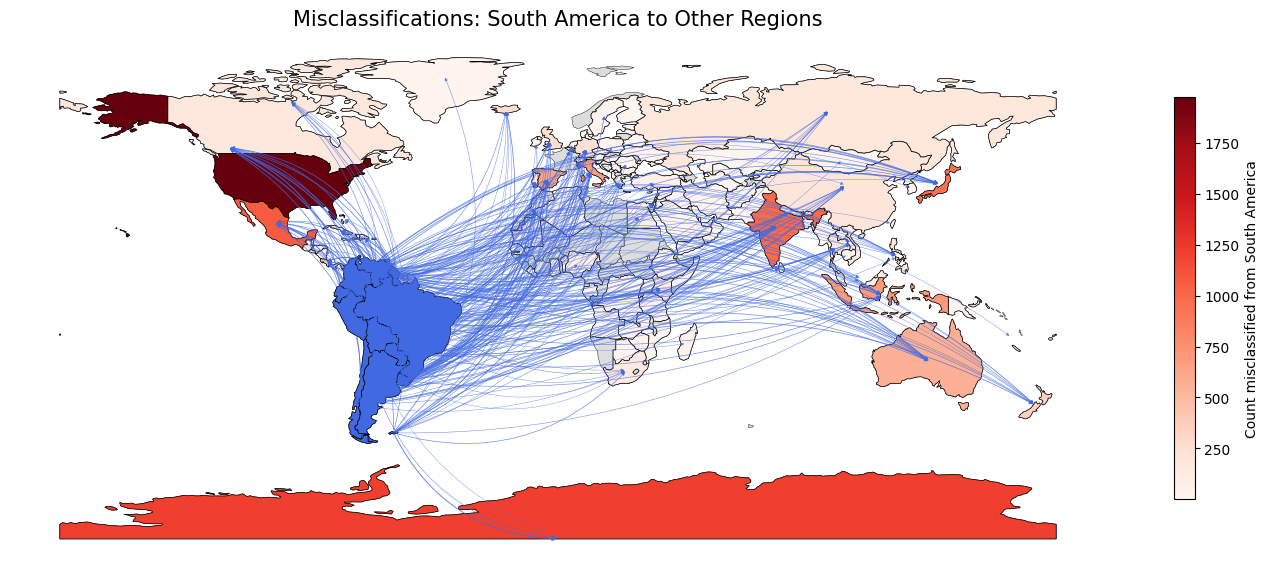}
    \caption{Mis-classification map : South America}
    \label{fig:815}
\end{figure*}

\begin{figure*}
    \centering
    \includegraphics[width=0.95\linewidth]{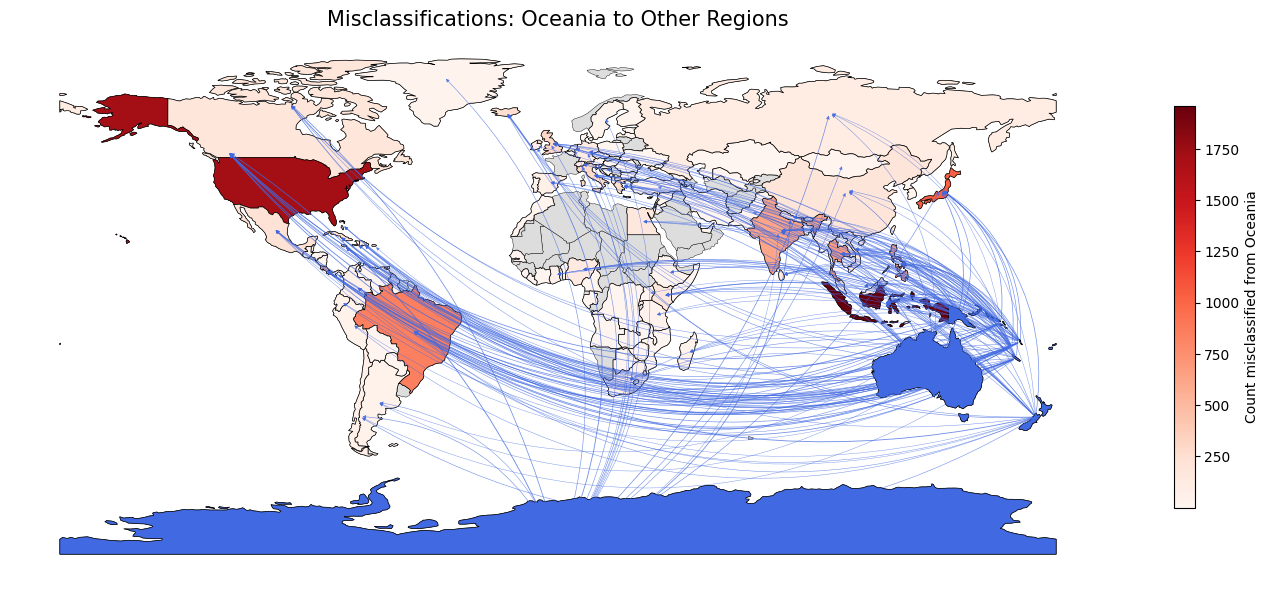}
    \caption{Mis-classification map : Oceania}
    \label{fig:816}
\end{figure*}

The mis-classifications from one region to countries outside the region can be seen fro each region in \autoref{fig:801} to \autoref{fig:816} respectively.

%%%%%%%%%%%%%%%%%%%%%%%%%%%%%%%%%%%%%%%%%%%%%%%%%%%%%%%%%%%%%%%%%%%%%%%%%%%%%%%%%%%%%%%%%%%%%%%%%%%%%%%%%%%%
\section{Country wise accuracies in each experimental setting}
\label{sec:appendix-Y}
The accuracies obtained over samples of each country through each experimental setup can be seen in \autoref{table:91} to \autoref{table:95}.
\begin{table}[!ht]
    \centering
    \begin{tabular}{cccc}
    \noalign{\hrule height 2pt}
    \rowcolor{purple2}\textbf{\tiny{Country name}} & \textbf{\tiny{Open-Ended}} & \textbf{\tiny{MCQs with}} & \textbf{\tiny{MCQs with}} \\
    \rowcolor{purple2}& & \textbf{\tiny{Similar choices}} & \textbf{\tiny{Random choices}} \\
    \noalign{\hrule height 2pt}
    \tiny{Afghanistan} & \tiny{41.33} & \tiny{68.90} & \tiny{81.56} \\
    \tiny{Albania} & \tiny{20.00} & \tiny{42.80} & \tiny{67.64} \\
    \tiny{Algeria} & \tiny{10.50} & \tiny{29.73} & \tiny{65.71} \\
    \tiny{Andorra} & \tiny{12.00} & \tiny{59.63} & \tiny{72.41} \\
    \tiny{Angola} & \tiny{4.67} & \tiny{48.07} & \tiny{58.83} \\
    \tiny{Anguilla} & \tiny{2.00} & \tiny{15.27} & \tiny{58.51} \\
    \tiny{Antarctica} & \tiny{34.83} & \tiny{84.80} & \tiny{83.57} \\
    \tiny{Antigua and Barbuda} & \tiny{7.67} & \tiny{31.67} & \tiny{70.64} \\
    \tiny{Argentina} & \tiny{30.67} & \tiny{84.17} & \tiny{71.39} \\
    \tiny{Armenia} & \tiny{42.33} & \tiny{66.23} & \tiny{80.07} \\
    \tiny{Aruba} & \tiny{17.67} & \tiny{55.67} & \tiny{78.96} \\
    \tiny{Australia} & \tiny{44.50} & \tiny{87.90} & \tiny{69.58} \\
    \tiny{Austria} & \tiny{18.83} & \tiny{42.13} & \tiny{80.69} \\
    \tiny{Azerbaijan} & \tiny{20.00} & \tiny{46.83} & \tiny{66.45} \\
    \tiny{Bahamas} & \tiny{24.83} & \tiny{69.47} & \tiny{78.13} \\
    \tiny{Bahrain} & \tiny{21.00} & \tiny{63.00} & \tiny{73.94} \\
    \tiny{Bangladesh} & \tiny{42.50} & \tiny{59.30} & \tiny{87.48} \\
    \tiny{Barbados} & \tiny{17.67} & \tiny{39.50} & \tiny{72.07} \\
    \tiny{Belarus} & \tiny{13.33} & \tiny{45.60} & \tiny{72.98} \\
    \tiny{Belgium} & \tiny{21.00} & \tiny{44.93} & \tiny{72.21} \\
    \tiny{Belize} & \tiny{11.67} & \tiny{59.13} & \tiny{68.49} \\
    \tiny{Benin} & \tiny{7.50} & \tiny{51.47} & \tiny{78.75} \\
    \tiny{Bermuda} & \tiny{20.67} & \tiny{62.63} & \tiny{67.61} \\
    \tiny{Bhutan} & \tiny{59.17} & \tiny{66.03} & \tiny{90.70} \\
    \tiny{Plurinational State of Bolivia} & \tiny{26.33} & \tiny{76.13} & \tiny{78.26} \\
    \tiny{Bonaire, Sint Eustatius and Saba} & \tiny{3.50} & \tiny{36.47} & \tiny{69.24} \\
    \tiny{Bosnia and Herzegovina} & \tiny{23.33} & \tiny{44.43} & \tiny{73.23} \\
    \tiny{Botswana} & \tiny{22.83} & \tiny{82.13} & \tiny{80.00} \\
    \tiny{Brazil} & \tiny{47.67} & \tiny{83.37} & \tiny{74.70} \\
    \tiny{Brunei Darussalam} & \tiny{8.67} & \tiny{21.73} & \tiny{48.78} \\
    \tiny{Bulgaria} & \tiny{25.33} & \tiny{46.47} & \tiny{77.12} \\
    \tiny{Burkina Faso} & \tiny{7.50} & \tiny{60.83} & \tiny{74.72} \\
    \tiny{Cabo Verde} & \tiny{10.17} & \tiny{67.23} & \tiny{55.22} \\
    \tiny{Cambodia} & \tiny{62.83} & \tiny{81.02} & \tiny{92.15} \\
    \tiny{Cameroon} & \tiny{4.67} & \tiny{67.20} & \tiny{70.02} \\
    \tiny{Canada} & \tiny{41.50} & \tiny{69.43} & \tiny{81.16} \\
    \tiny{Cayman Islands} & \tiny{6.67} & \tiny{28.07} & \tiny{68.78} \\
    \tiny{Central African Republic} & \tiny{0.83} & \tiny{16.67} & \tiny{50.21} \\
    \tiny{Chile} & \tiny{20.83} & \tiny{65.90} & \tiny{67.78} \\
    \tiny{China} & \tiny{58.83} & \tiny{78.73} & \tiny{81.48} \\
    \tiny{Colombia} & \tiny{23.83} & \tiny{75.73} & \tiny{69.25} \\
    \tiny{Democratic Republic of  Congo} & \tiny{6.83} & \tiny{40.70} & \tiny{56.60} \\
    \tiny{Cook Islands} & \tiny{3.83} & \tiny{22.23} & \tiny{68.28} \\
    \noalign{\hrule height 2pt}
    \end{tabular}
    \caption{Country wise accuracies through various experimental settings : Part 1/5}
    \label{table:91}
\end{table}

\begin{table}[!ht]
    \centering
    \begin{tabular}{cccc}
    \noalign{\hrule height 2pt}
    \rowcolor{purple2}\textbf{\tiny{Country name}} & \textbf{\tiny{Open-Ended}} & \textbf{\tiny{MCQs with}} & \textbf{\tiny{MCQs with}} \\
    \rowcolor{purple2}& & \textbf{\tiny{Similar choices}} & \textbf{\tiny{Random choices}} \\
    \noalign{\hrule height 2pt}
    \tiny{Costa Rica} & \tiny{26.00} & \tiny{73.23} & \tiny{72.16} \\
    \tiny{Croatia} & \tiny{47.83} & \tiny{72.83} & \tiny{83.92} \\
    \tiny{Cuba} & \tiny{47.50} & \tiny{76.83} & \tiny{77.92} \\
    \tiny{Curaçao} & \tiny{20.83} & \tiny{61.07} & \tiny{80.96} \\
    \tiny{Cyprus} & \tiny{13.67} & \tiny{59.33} & \tiny{69.19} \\
    \tiny{Czechia} & \tiny{40.50} & \tiny{66.07} & \tiny{83.90} \\
    \tiny{Côte d'Ivoire} & \tiny{13.33} & 
    \tiny{60.00} &
    \tiny{71.47} \\
    \tiny{Denmark} & \tiny{32.50} & \tiny{66.93} & \tiny{78.54} \\
    \tiny{Dominica} & \tiny{15.17} & \tiny{61.17} & \tiny{67.04} \\
    \tiny{Dominican Republic} & \tiny{15.00} & \tiny{56.37} & \tiny{70.43} \\
    \tiny{Ecuador} & \tiny{21.50} & \tiny{76.10} & \tiny{73.05} \\
    \tiny{Egypt} & \tiny{60.50} & \tiny{77.07} & \tiny{83.84} \\
    \tiny{El Salvador} & \tiny{4.83} & \tiny{65.93} & \tiny{63.40} \\
    \tiny{Estonia} & \tiny{21.83} & \tiny{43.90} & \tiny{70.30} \\
    \tiny{Eswatini} & \tiny{0.50} & \tiny{28.70} & \tiny{53.07} \\
    \tiny{Ethiopia} & \tiny{41.00} & \tiny{80.93} & \tiny{80.24} \\
    \tiny{Falkland Islands} & \tiny{8.83} & \tiny{92.13} & \tiny{90.35} \\
    \tiny{Faroe Islands} & \tiny{30.33} & \tiny{71.30} & \tiny{90.66} \\
    \tiny{Fiji} & \tiny{22.83} & \tiny{64.10} & \tiny{76.93} \\
    \tiny{Finland} & \tiny{32.33} & \tiny{67.80} & \tiny{76.31} \\
    \tiny{France} & \tiny{40.83} & \tiny{73.77} & \tiny{83.70} \\
    \tiny{French Guiana} & \tiny{3.00} & \tiny{64.73} & \tiny{53.93} \\
    \tiny{French Polynesia} & \tiny{24.67} & \tiny{81.90} & \tiny{83.97} \\
    \tiny{Gabon} & \tiny{5.33} & \tiny{56.00} & \tiny{66.67} \\
    \tiny{Gambia} & \tiny{3.33} & \tiny{41.40} & \tiny{53.16} \\
    \tiny{Georgia} & \tiny{32.00} & \tiny{71.40} & \tiny{83.37} \\
    \tiny{Germany} & \tiny{54.83} & \tiny{71.30} & \tiny{87.54} \\
    \tiny{Ghana} & \tiny{26.33} & \tiny{70.53} & \tiny{67.80} \\
    \tiny{Gibraltar} & \tiny{19.00} & \tiny{62.27} & \tiny{79.48} \\
    \tiny{Greece} & \tiny{66.67} & \tiny{91.00} & \tiny{91.06} \\
    \tiny{Greenland} & \tiny{27.00} & \tiny{65.43} & \tiny{84.90} \\
    \tiny{Grenada} & \tiny{3.50} & \tiny{37.77} & \tiny{63.75} \\
    \tiny{Guadeloupe} & \tiny{1.50} & \tiny{48.80} & \tiny{71.09} \\
    \tiny{Guam} & \tiny{11.33} & \tiny{70.43} & \tiny{55.57} \\
    \tiny{Guatemala} & \tiny{19.67} & \tiny{75.50} & \tiny{74.38} \\
    \tiny{Guernsey} & \tiny{1.83} & \tiny{66.50} & \tiny{81.14} \\
    \tiny{Guyana} & \tiny{8.83} & \tiny{52.33} & \tiny{52.66} \\
    \tiny{Haiti} & \tiny{27.83} & \tiny{71.23} & \tiny{65.98} \\
    \tiny{Vatican City State} & \tiny{8.67} & \tiny{43.77} & \tiny{74.31} \\
    \tiny{Honduras} & \tiny{4.67} & \tiny{64.13} & \tiny{66.98} \\
    \tiny{Hong Kong} & \tiny{22.67} & \tiny{65.23} & \tiny{86.70} \\
    \tiny{Hungary} & \tiny{24.83} & \tiny{49.00} & \tiny{78.44} \\
    \tiny{Iceland} & \tiny{69.00} & \tiny{82.27} & \tiny{89.04} \\
    \noalign{\hrule height 2pt}
    \end{tabular}
    \caption{Region wise accuracies through various experimental settings : Part 2/5}
    \label{table:92}
\end{table}

\begin{table}[!ht]
    \centering
    \begin{tabular}{cccc}
    \noalign{\hrule height 2pt}
    \rowcolor{purple2}\textbf{\tiny{Country name}} & \textbf{\tiny{Open-Ended}} & \textbf{\tiny{MCQs with}} & \textbf{\tiny{MCQs with}} \\
    \rowcolor{purple2}& & \textbf{\tiny{Similar choices}} & \textbf{\tiny{Random choices}} \\
    \noalign{\hrule height 2pt}
    \tiny{India} & \tiny{78.33} & \tiny{90.03} & \tiny{90.10} \\
    \tiny{Indonesia} & \tiny{48.83} & \tiny{67.76} & \tiny{84.97} \\
    \tiny{Iran} & \tiny{50.83} & \tiny{70.40} & \tiny{83.27} \\
    \tiny{Iraq} & \tiny{28.67} & \tiny{60.60} & \tiny{76.84} \\
    \tiny{Ireland} & \tiny{48.33} & \tiny{74.57} & \tiny{87.63} \\
    \tiny{Isle of Man} & \tiny{6.17} & \tiny{52.03} & \tiny{77.91} \\
    \tiny{Israel} & \tiny{35.67} & \tiny{76.33} & \tiny{73.99} \\
    \tiny{Italy} & \tiny{60.00} & \tiny{82.30} & \tiny{85.40} \\
    \tiny{Jamaica} & \tiny{28.17} & \tiny{60.20} & \tiny{70.58} \\
    \tiny{Japan} & \tiny{81.17} & \tiny{88.92} & \tiny{91.75} \\
    \tiny{Jersey} & \tiny{3.67} & \tiny{50.37} & \tiny{71.69} \\
    \tiny{Jordan} & \tiny{44.00} & \tiny{79.03} & \tiny{89.04} \\
    \tiny{Kazakhstan} & \tiny{18.33} & \tiny{44.73} & \tiny{77.73} \\
    \tiny{Kenya} & \tiny{56.00} & \tiny{88.57} & \tiny{88.15} \\
    \tiny{North Korea} & \tiny{47.33} & \tiny{25.64}& \tiny{81.26}\\
    \tiny{South Korea} & \tiny{47.83} & \tiny{67.23} & \tiny{79.90} \\
    \tiny{Kuwait} & \tiny{12.83} & \tiny{52.30} & \tiny{68.70} \\
    \tiny{Kyrgyzstan} & \tiny{20.17} & \tiny{37.30} & \tiny{69.48} \\
    \tiny{Laos} & \tiny{26.50} & \tiny{38.53} & \tiny{80.25}\\
    \tiny{Latvia} & \tiny{17.00} & \tiny{41.63} & \tiny{72.47} \\
    \tiny{Lebanon} & \tiny{27.00} & \tiny{73.63} & \tiny{78.09} \\
    \tiny{Liberia} & \tiny{9.33} & \tiny{50.97} & \tiny{65.37} \\
    \tiny{Libya} & \tiny{6.67} & \tiny{22.87} & \tiny{73.10} \\
    \tiny{Liechtenstein} & \tiny{6.17} & \tiny{34.03} & \tiny{72.29} \\
    \tiny{Lithuania} & \tiny{24.00} & \tiny{54.43} & \tiny{74.40} \\
    \tiny{Luxembourg} & \tiny{13.33} & \tiny{21.90} & \tiny{62.29} \\
    \tiny{Macao} & \tiny{17.00} & \tiny{66.42} & \tiny{85.38} \\
    \tiny{Madagascar} & \tiny{24.17} & \tiny{81.20} & \tiny{65.40} \\
    \tiny{Malawi} & \tiny{8.33} & \tiny{54.80} & \tiny{66.39} \\
    \tiny{Malaysia} & \tiny{28.33} & \tiny{73.28} & \tiny{83.65} \\
    \tiny{Maldives} & \tiny{39.33} & \tiny{80.20} & \tiny{82.08} \\
    \tiny{Mali} & \tiny{13.83} & \tiny{65.43} & \tiny{80.11} \\
    \tiny{Malta} & \tiny{47.67} & \tiny{79.57} & \tiny{90.95} \\
    \tiny{Martinique} & \tiny{4.33} & \tiny{53.60} & \tiny{72.85} \\
    \tiny{Mauritania} & \tiny{12.00} & \tiny{76.77} & \tiny{80.28} \\
    \tiny{Mauritius} & \tiny{38.33} & \tiny{92.00} & \tiny{79.52} \\
    \tiny{Mexico} & \tiny{53.17} & \tiny{79.77} & \tiny{79.69} \\
    \tiny{Moldova} & \tiny{7.67} & \tiny{35.23} & \tiny{63.57} \\
    \tiny{Monaco} & \tiny{30.17} & \tiny{54.83} & \tiny{69.69} \\
    \tiny{Mongolia} & \tiny{50.83} & \tiny{82.41} & \tiny{81.39} \\
    \tiny{Montenegro} & \tiny{22.17} & \tiny{44.37} & \tiny{81.00} \\
    \tiny{Morocco} & \tiny{67.83} & \tiny{85.40} & \tiny{93.75} \\
    \tiny{Mozambique} & \tiny{5.17} & \tiny{66.57} & \tiny{63.78} \\
    \tiny{Myanmar} & \tiny{61.50} & \tiny{76.56} & \tiny{92.62} \\
    \noalign{\hrule height 2pt}
    \end{tabular}
    \caption{Region wise accuracies through various experimental settings : Part 3/5}
    \label{table:93}
\end{table}

\begin{table}[!ht]
    \centering
    \begin{tabular}{cccc}
    \noalign{\hrule height 2pt}
    \rowcolor{purple2}\textbf{\tiny{Country name}} & \textbf{\tiny{Open-Ended}} & \textbf{\tiny{MCQs with}} & \textbf{\tiny{MCQs with}} \\
    \rowcolor{purple2}& & \textbf{\tiny{Similar choices}} & \textbf{\tiny{Random choices}} \\
    \noalign{\hrule height 2pt}
    \tiny{Namibia} & \tiny{0.00} & \tiny{83.40} & \tiny{85.35} \\
    \tiny{Nepal} & \tiny{65.00} & \tiny{72.53} & \tiny{89.72} \\
    \tiny{Netherlands} & \tiny{46.00} & \tiny{74.63} & \tiny{86.86} \\
    \tiny{New Caledonia} & \tiny{7.50} & \tiny{55.03} & \tiny{64.98} \\
    \tiny{New Zealand} & \tiny{53.83} & \tiny{76.40} & \tiny{82.58} \\
    \tiny{Nicaragua} & \tiny{6.83} & \tiny{69.87} & \tiny{69.64} \\
    \tiny{Nigeria} & \tiny{47.33} & \tiny{79.13} & \tiny{73.78} \\
    \tiny{North Macedonia} & \tiny{10.17} & \tiny{44.27} & \tiny{74.44} \\
    \tiny{Norway} & \tiny{32.50} & \tiny{48.17} & \tiny{79.45} \\
    \tiny{Oman} & \tiny{31.67} & \tiny{71.40} & \tiny{77.59} \\
    \tiny{Pakistan} & \tiny{30.33} & \tiny{53.57} & \tiny{79.32} \\
    \tiny{Palau} & \tiny{15.83} & \tiny{71.23} & \tiny{71.97} \\
    \tiny{Palestine, State of} & \tiny{9.00} & \tiny{73.53} & \tiny{83.59} \\
    \tiny{Panama} & \tiny{4.33} & \tiny{80.17} & \tiny{60.86} \\
    \tiny{Papua New Guinea} & \tiny{13.50} & \tiny{61.87} & \tiny{63.38} \\
    \tiny{Paraguay} & \tiny{6.17} & \tiny{52.23} & \tiny{54.29} \\
    \tiny{Peru} & \tiny{54.83} & \tiny{85.73} & \tiny{83.61} \\
    \tiny{Philippines} & \tiny{43.67} & \tiny{74.82} & \tiny{85.94} \\
    \tiny{Poland} & \tiny{28.83} & \tiny{62.00} & \tiny{79.17} \\
    \tiny{Portugal} & \tiny{43.50} & \tiny{58.60} & \tiny{84.39} \\
    \tiny{Puerto Rico} & \tiny{16.67} & \tiny{68.97} & \tiny{72.52} \\
    \tiny{Qatar} & \tiny{19.50} & \tiny{56.63} & \tiny{66.04} \\
    \tiny{Romania} & \tiny{31.50} & \tiny{56.43} & \tiny{79.02} \\
    \tiny{Russian Federation} & \tiny{52.67} & \tiny{73.13} & \tiny{77.18} \\
    \tiny{Rwanda} & \tiny{29.50} & \tiny{71.73} & \tiny{73.72} \\
    \tiny{Réunion} & \tiny{5.33} & \tiny{90.87} & \tiny{69.21} \\
    \tiny{Saint Helena, Ascension} & \tiny{3.33} & \tiny{71.40} & \tiny{57.44} \\
    \tiny{and Tristan da Cunha} & \tiny{} & \tiny{} & \tiny{} \\
    \tiny{Saint Kitts and Nevis} & \tiny{14.17} & \tiny{41.23} & \tiny{64.61} \\
    \tiny{Saint Lucia} & \tiny{16.83} & \tiny{61.40} & \tiny{79.33} \\
    \tiny{Saint Martin (French)} & \tiny{4.00} & \tiny{45.43} & \tiny{69.48} \\
    \tiny{Samoa} & \tiny{23.33} & \tiny{68.43} & \tiny{71.19} \\
    \tiny{San Marino} & \tiny{10.17} & \tiny{35.00} & \tiny{54.01} \\
    \tiny{Saudi Arabia} & \tiny{26.00} & \tiny{65.53} & \tiny{74.69} \\
    \tiny{Senegal} & \tiny{21.83} & \tiny{78.73} & \tiny{78.20} \\
    \tiny{Serbia} & \tiny{24.33} & \tiny{58.70} & \tiny{79.14} \\
    \tiny{Seychelles} & \tiny{26.33} & \tiny{92.87} & \tiny{76.83} \\
    \tiny{Sierra Leone} & \tiny{8.83} & \tiny{56.53} & \tiny{75.23} \\
    \tiny{Singapore} & \tiny{51.33} & \tiny{74.91} & \tiny{80.15} \\
    \tiny{Saint Martin (Dutch)} & \tiny{7.17} & \tiny{50.77} & \tiny{75.14} \\
    \tiny{Slovakia} & \tiny{12.33} & \tiny{32.33} & \tiny{67.41} \\
    \tiny{Slovenia} & \tiny{24.00} & \tiny{53.40} & \tiny{75.09} \\
    \tiny{Solomon Islands} & \tiny{3.33} & \tiny{22.53} & \tiny{69.22} \\  
    \tiny{Somalia} & \tiny{24.67} & \tiny{75.30} & \tiny{78.46} \\      
    \noalign{\hrule height 2pt}
    \end{tabular}
    \caption{Region wise accuracies through various experimental settings : Part 4/5}
    \label{table:94}
\end{table}

\begin{table}[!ht]
    \centering
    \begin{tabular}{cccc}
    \noalign{\hrule height 2pt}
    \rowcolor{purple2}\textbf{\tiny{Country name}} & \textbf{\tiny{Open-Ended}} & \textbf{\tiny{MCQs with}} & \textbf{\tiny{MCQs with}} \\
    \rowcolor{purple2}& & \textbf{\tiny{Similar choices}} & \textbf{\tiny{Random choices}} \\
    \noalign{\hrule height 2pt}
    \tiny{South Africa} & \tiny{38.50} & \tiny{94.43} & \tiny{82.91} \\
    \tiny{South Georgia and the} & \tiny{7.17} & \tiny{80.70} & \tiny{77.99} \\
    \tiny{South Sandwich Islands} & & & \\
    \tiny{South Sudan} & \tiny{25.83} & \tiny{65.83} & \tiny{82.31} \\
    \tiny{Spain} & \tiny{51.00} & \tiny{83.13} & \tiny{84.71} \\
    \tiny{Sri Lanka} & \tiny{37.00} & \tiny{61.40} & \tiny{82.72} \\
    \tiny{Sudan} & \tiny{25.33} & \tiny{70.63} & \tiny{81.25} \\
    \tiny{Svalbard and Jan Mayen} & \tiny{0.00} & \tiny{74.13} & \tiny{89.45} \\
    \tiny{Sweden} & \tiny{35.50} & \tiny{54.63} & \tiny{81.22} \\
    \tiny{Switzerland} & \tiny{42.17} & \tiny{62.53} & \tiny{76.40} \\
    \tiny{Syrian Arab Republic} & \tiny{13.00} & \tiny{51.63} & \tiny{64.82} \\
    \tiny{Taiwan, Province of China} & \tiny{23.00} & \tiny{51.01} & \tiny{80.16} \\
    \tiny{Tajikistan} & \tiny{10.83} & \tiny{44.43} & \tiny{81.04} \\
    \tiny{Tanzania, United Republic of} & \tiny{24.83} & \tiny{84.37} & \tiny{84.89} \\
    \tiny{Thailand} & \tiny{64.17} & \tiny{84.49} & \tiny{89.08} \\
    \tiny{Timor-Leste} & \tiny{7.83} & \tiny{41.77} & \tiny{69.67} \\
    \tiny{Togo} & \tiny{2.33} & \tiny{31.67} & \tiny{65.98} \\
    \tiny{Tonga} & \tiny{1.33} & \tiny{19.60} & \tiny{44.73} \\
    \tiny{Trinidad and Tobago} & \tiny{8.00} & \tiny{56.23} & \tiny{53.62} \\
    \tiny{Tunisia} & \tiny{20.33} & \tiny{40.00} & \tiny{75.53} \\
    \tiny{Turkmenistan} & \tiny{22.67} & \tiny{48.73} & \tiny{82.83} \\
    \tiny{Türkiye} & \tiny{56.33} & \tiny{86.10} & \tiny{92.24} \\
    \tiny{Uganda} & \tiny{26.83} & \tiny{79.90} & \tiny{80.27} \\
    \tiny{Ukraine} & \tiny{22.83} & \tiny{67.63} & \tiny{72.82} \\
    \tiny{United Arab Emirates} & \tiny{53.00} & \tiny{85.30} & \tiny{85.30} \\
    \tiny{United Kingdom} & \tiny{50.17} & \tiny{92.17} & \tiny{89.05} \\
    \tiny{United States} & \tiny{67.17} & \tiny{91.03} & \tiny{87.76} \\
    \tiny{Uruguay} & \tiny{14.17} & \tiny{46.33} & \tiny{61.10} \\
    \tiny{Uzbekistan} & \tiny{47.17} & \tiny{68.63} & \tiny{83.07} \\
    \tiny{Vanuatu} & \tiny{5.50} & \tiny{18.00} & \tiny{57.04} \\
    \tiny{Venezuela, Bolivarian Republic of} & \tiny{11.17} & \tiny{57.63} & \tiny{53.41} \\
    \tiny{Viet Nam} & \tiny{55.50} & \tiny{78.74} & \tiny{89.77} \\
    \tiny{Virgin Islands, British} & \tiny{6.83} & \tiny{38.00} & \tiny{79.60} \\
    \tiny{Virgin Islands, U.S.} & \tiny{9.67} & \tiny{46.73} & \tiny{81.72} \\
    \tiny{Kosovo} & \tiny{6.50} & \tiny{28.70} & \tiny{65.53} \\
    \tiny{Yemen} & \tiny{27.17} & \tiny{69.80} & \tiny{76.46} \\
    \tiny{Zambia} & \tiny{9.50} & \tiny{54.80} & \tiny{73.29} \\
    \tiny{Zimbabwe} & \tiny{11.67} & \tiny{71.03} & \tiny{76.05} \\
    \tiny{\AA land Islands} & \tiny{0.17} & \tiny{29.00} & \tiny{62.02} \\
    \noalign{\hrule height 2pt}
    \textbf{\tiny{Overall}}         & \textbf{\tiny{25.14}} & \textbf{\tiny{61.92}} & \textbf{\tiny{75.06}} \\
    \noalign{\hrule height 2pt}
    \end{tabular}
    \caption{Region wise accuracies through various experimental settings : Part 5/5}
    \label{table:95}
\end{table}
%%%%%%%%%%%%%%%%%%%%%%%%%%%%%%%%%%%%%%%%%%%%%%%%%%%%%%%%%%%%%%%%%%%%%%%%%%%%%%%%%%%%%%%%%%%%%%%%%%%%%%%%%%%%
%%%------------------------------------------------------------------------------------------------------%%%
\begin{figure*}
    \centering
    \includegraphics[width=1\linewidth]{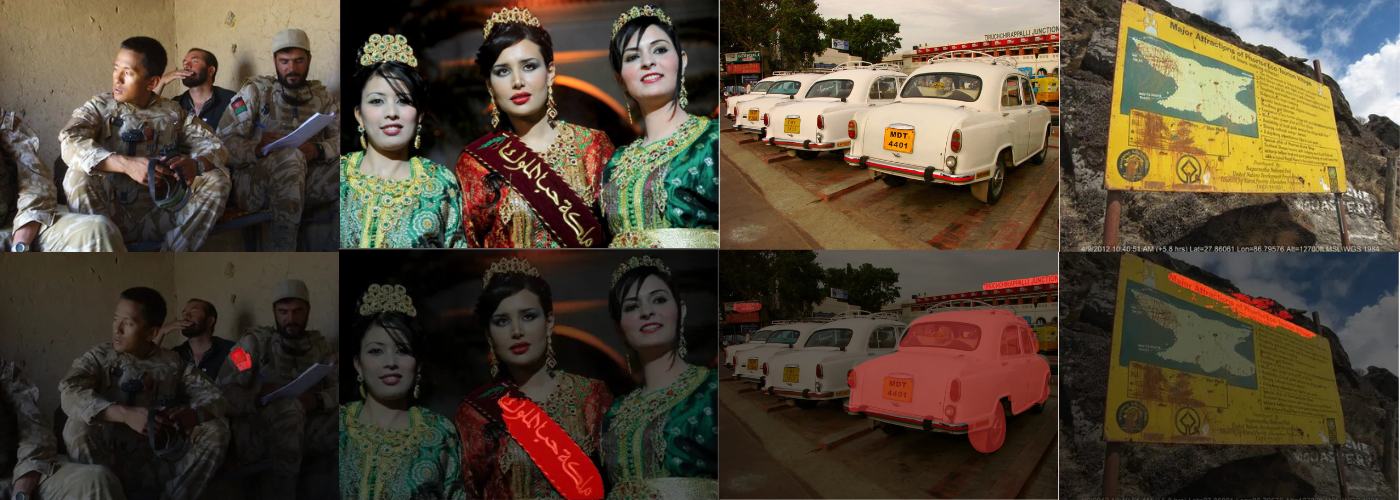}
    \caption{Examples from ours (1st,4th) as well as other works : GIMMICK (2nd), CVQA (3rd) : The 1st and 4th image have the key features required for classifying the image accurately, occupying a tiny portion of the image making it relatively difficult i.e the flag patch in image 1 ,and name of mountain in image 4's signboard. While, in Image 2 and image 3 , the key features i.e the text on attire or the (car, city name signboard, multilingual texts on left) make the samples relatively easier to classify}
    \label{fig-label}
\end{figure*}
\begin{figure}
    \centering
    \includegraphics[width=1\linewidth]{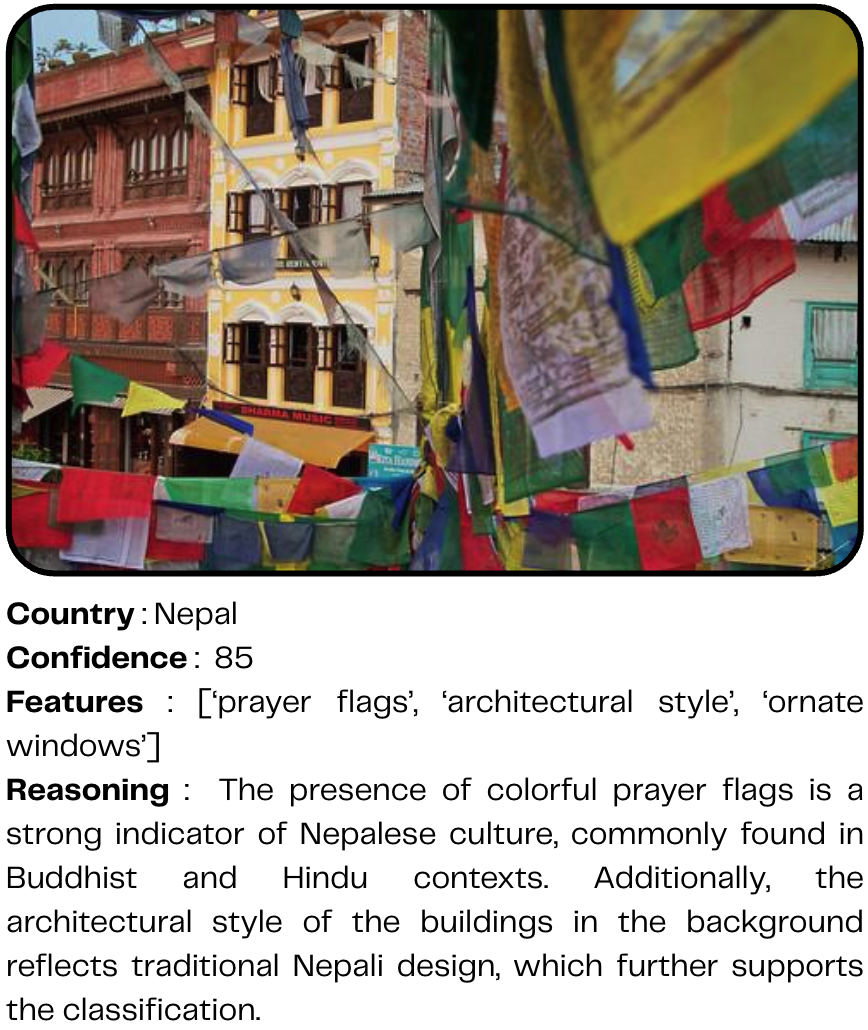}
    \caption{A sample from our dataset and its corresponding response (GPT-4o-Mini)}
    \label{figure:2}
\end{figure}
%%%%%%%%%%%%%%%%%%%%%%%%%%%%%%%%%%%%%%%%%%%%%%%%%%%%%%%%%%%%%%%%%%%%%%%%%%%%%%%%%%%%%%%%%%%%%%%%%%%%%%%%%%%%
\end{document}